%% file: arxiv_preprint.tex
\documentclass[journal]{IEEEtran}
\usepackage[utf8]{inputenc}
\usepackage{cite}
\usepackage{amsmath,amssymb,amsfonts,mathtools}
\usepackage{bm}
\usepackage{algorithm, algorithmic}
\usepackage{booktabs, multirow, makecell, tabularray}
\usepackage{graphicx}
\usepackage{flushend}
\usepackage{textcomp}
\usepackage{xcolor}
\usepackage{hyperref}
\usepackage{rotating}
\usepackage{mleftright}
\usepackage{wrapfig}
\usepackage{float}
\usepackage{mdwlist}
\usepackage{subcaption}
\usepackage{makecell}
\usepackage{orcidlink}

\input{figures/colors.tex}

\DeclareMathOperator{\relu}{ReLu}

\DeclareMathOperator{\mean}{mean}
\DeclareMathOperator{\pnorm}{\mathnormal{p}-norm}
\DeclareMathOperator{\lse}{lse}

\mleftright 

\usepackage{array}
\usepackage{eqparbox}

\title{Malicious Internet Entity Detection Using Local Graph Inference}

\author{
Šimon Mandlík\,\orcidlink{0000-0002-8776-2854},
Tomáš Pevný\,\orcidlink{0000-0002-5768-9713},
Václav Šmídl\,\orcidlink{0000-0003-3027-6174},
Lukáš Bajer\,\orcidlink{0000-0002-9402-6417}
\thanks{Šimon Mandlík, Tomáš Pevný and Václav Šmídl are with Artificial Intelligence Center, Department of Computer Science, Czech Technical University, Prague (email: mandlsim@fel.cvut.cz; pevnytom@fel.cvut.cz; smidlva1@fel.cvut.cz).}
\thanks{Lukáš Bajer is with Cisco Systems, Prague, Czech Republic (email: lubajer@cisco.com).}
}

\begin{document}

\maketitle

\textit{This document is a preprint of \cite{Mandlik2024}.}
\newline

\begin{abstract}
Detection of malicious behavior in a large network is a challenging problem for machine learning in computer security since it requires a model with high expressive power and scalable inference. Existing solutions struggle to achieve this feat---current cybersec-tailored approaches are still limited in expressivity, and methods successful in other domains do not scale well for large volumes of data rendering frequent retraining impossible. 
This work proposes a new perspective for learning from graph data that is modeling network entity interactions as a large heterogeneous graph. High expressivity of the method is achieved with a neural network architecture HMILnet that naturally models this type of data and provides theoretical guarantees. The scalability is achieved by pursuing local graph inference in a streamlined neighborhood subgraph, i.e., classifying individual vertices and their neighborhood as independent samples. Our experiments exhibit improvements over the state-of-the-art Probabilistic Threat Propagation (PTP) algorithm, show a threefold accuracy improvement when additional data is used, which is not possible with the PTP algorithm, and demonstrate the generalization capabilities of the method to new, previously unseen entities.
\end{abstract}

\begin{IEEEkeywords}
Multiple Instance Learning, Hierarchical Multiple Instance Learning, Graph Deep Learning, Computer
Security, Threat Detection, Communication Networks
\end{IEEEkeywords}

\section{Introduction}%
\label{sec:introduction}
Detection of malicious activities in a network is a critical concern in cybersecurity. A
\emph{denylist} containing entities (second-level domains, IP addresses, emails, etc.) known for being
involved in malicious activities is important either for basic blacklisting or as a source of labels
for more sophisticated solutions based on machine learning. The major problem is that denylist
quickly ages with time and therefore needs to be frequently updated, mostly with the help of machine
learning to decrease the cost of labeling.

The most studied approach trains a classifier predicting \emph{maliciousness} of each entity
independently of others based on manually-designed features using denylist as a source of labels.
For example, one can detect malicious domains with a model operating on a vector of features
extracted from their URLs \cite{Bartos2016}. This \emph{feature-based approach} is extensively
researched~\cite{zhang2008, Antonakakis2010BuildingAD, bilge2011, Bartos2016}, despite its
weaknesses like limited expressiveness, the requirement of expert domain knowledge to design features,
and possibly deteriorating performance over time due to \emph{concept drift}\cite{Pendlebury2018}.

An alternative is to model interactions of entities, which is most naturally modeled 
as a \emph{graph}. \emph{Graph-based} methods in cybersecurity~\cite{Carter2014a,coskun2010friends,philips2012detecting} 
assume that malicious activities are localized in the graph~\cite{Yu2010,collins2007using}, 
forming communities with sharp boundaries. Algorithms propagate maliciousness from denylisted entities
to others using fixed formulas~\cite{oprea2015, Rezvani, PlumeWalk} with few or no hyper-parameters. This
is appealing due to the seeming absence of training, but it limits the fine-tuning of the algorithm to new domains.
To the best of our knowledge, these classical graph algorithms cannot utilize features on edges and vertices 
other than scalars.

Although the feature-based and graph-based paradigms seem complementary, one can identify several
similarities. First, graph-based algorithms implicitly use features in the form of the graph (neighborhood).
For example, Refs.~\cite{Carter2013, Carter2014a} estimate maliciousness of a vertex while using maliciousness 
of neighboring vertices and edge weights. Secondly, authors of a new graph algorithm frequently select one 
algorithm out of a set of candidates, which can be viewed as a form of training. In defense of graph-based
algorithms, their features frequently consist only of a scalar value expressing \emph{maliciousness}
of vertices and scalar weights of edges, which makes them, to some extent, robust to changes of
application domain.

We advocate a third approach (called \emph{neighborhood-based}) that bridges the gap between the
feature-based and graph-based methods~\cite{Hamilton2017, Zhou2018a, Wu2019, Busch2021, Herath2022,
Wei2022, Zhu2022}. Methods following this approach utilize features available for vertices and edges
from a small local neighborhood. By changing the size of the neighborhood and adding or removing
information in vertices and edges, neighborhood-based methods can approach the behavior of one of the
two aforementioned approaches without modifications of the core algorithm.

For the purpose of this work, we use the term `graph inference' for a process of estimating a scalar
value of maliciousness for the entities of interest using a graph structure as the main source of
knowledge.

\begin{figure*}[ht] 
    \centering
    \includegraphics[width=\textwidth]{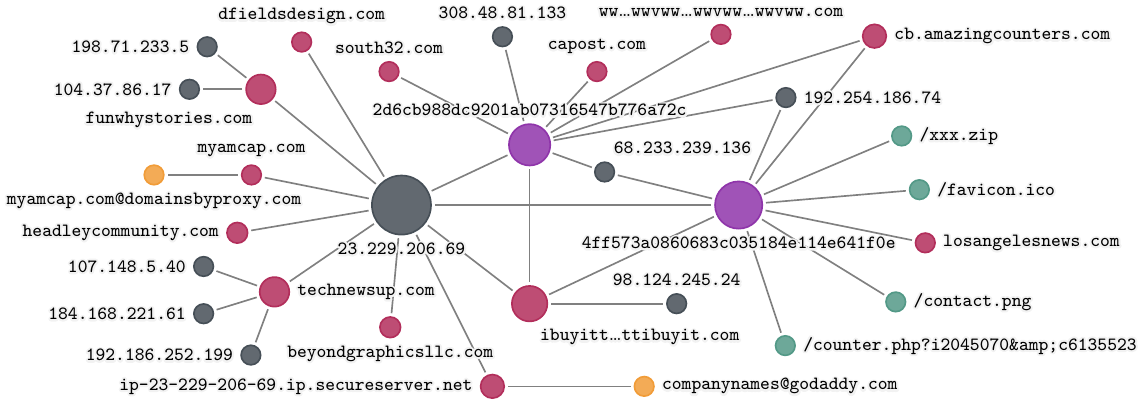}
    \caption{An example of a network graph of binaries (represented by SHA hashes) in
        {\color{shacolor}\textbf{purple}}, second-level domains in {\color{domaincolor}\textbf{red}}, URL paths in
        {\color{pathcolor}\textbf{green}}, IP addresses in {\color{ipcolor}\textbf{gray}}, and emails in
        {\color{emailcolor}\textbf{yellow}}. Edges represent an interaction, for example, when a binary has
        contacted a domain hosted on an IP address or when a domain is registered with an email
        address. Sequences of many repeated characters in URL names were shortened using ellipsis.
    The example is taken from~\url{https://www.threatcrowd.org/}, and the graph was symmetrized for
demonstration purposes. Best viewed in color.}%
    \label{fig:threatcrowd}
\end{figure*}

This work presents a new neighborhood-based method for graph inference based on Hierarchical
Multiple Instance Learning (HMIL) \cite{Dietterich1997,Maron1998, Amores2013, Foulds2010,
Pevny2016,Pevny2017a,Edwards2016, Pevny2019} tailored to the needs of the cybersecurity domain. 
Overall, we make the following contributions:

\begin{itemize}
\item We describe the new algorithm in detail and discuss its homogeneous and
    heterogeneous variant.
\item We explain the fundamental differences to the Probabilistic Threat Propagation (PTP)
    algorithm\cite{Carter2013, Carter2014a} and Graph neural network
    methods~\cite{Bronstein2016,Kipf2016,Hamilton2017} and their heterogeneous
    extensions~\cite{Chang2015, Gui2016, Fu2017, Shi2017, Dong2017, Zhang2018, Sun2018, Fan2019,
    Wang2019}.
\item We show how the algorithm can be used to learn from elementary network observations in the
    form of multiple (heterogeneous) binary relations. 
\item The approach is first compared to the PTP algorithm~\cite{Carter2014a} in a limited setting
    (imposed by PTP) of modeling interactions between network clients and second-level domains. 
\item Then the generality of the proposed scheme is demonstrated on a task of detection of
    second-level domains involved in malicious activities, leveraging relations between a large number of
    entities in the Internet  (binaries, IP addresses, TLS certificates and WHOIS registries). To
    the best of our knowledge, this is the first work modeling interaction of such heterogeneous
    entities on this scale in this domain.
 \end{itemize}

The next Section~\ref{sec:data_and_problem_description} defines the problem and describes the nature of
the data available on the input. Section~\ref{sec:related_work} discusses relevant work on graphical
models and reviews advances in the (Hierarchical) Multiple Instance Learning field.
Section~\ref{sec:hmilnet_based_graph_inference} describes the graph processing approach based on
HMIL and Section~\ref{sec:mapping_the_internet} shows an extension of the method to heterogeneous
graphs, resulting in a general method for mapping the Internet using only several binary relations
as input. Finally, Section~\ref{sec:experiments} shows the experimental comparison and
Section~\ref{sec:conclusion} concludes the work and discusses possible future research directions.
In the Supplementary material, we provide a list of important symbols, full experimental results, and dataset details.

\section{Data and problem description}%
\label{sec:data_and_problem_description}
The goal of this work is to estimate the probability that any network entity (like a second-level domain, IP address, or email) is involved in malicious
activity. The denylist consisting of known malicious entities is used as a source of
knowledge, however, it is almost surely incomplete. Even though the prior probability that a random
non-denylisted entity is clean is high, the task is best solved as a \emph{positive-unlabeled
problem}~\cite{Blanchard2010}, considering denylisted entities as malicious and non-denylisted
entities as `unknown'. For evaluation purposes, we consider all non-denylisted entities
benign.

We estimate the maliciousness probability by observing raw interactions in the form of binary
relations involving entities in the Internet and the knowledge about their involvement in unwanted
actions (which is always incomplete). Binary relations are naturally modeled as a
\textit{heterogeneous} graph, where vertices correspond to Internet entities (second-level domains, IP addresses,
emails, clients, processes, etc.) and edges to interactions between them. An illustrative example is
shown in Figure~\ref{fig:threatcrowd}.

Since \cite{Blanchard2010} lists conditions for solving positive-unlabeled problems as a supervised
task, the problem can be viewed as a series of independent binary classifications of individual
vertices of interest, in which case the heterogeneous graph and denylist are an abstraction of
relational database.

The problem as defined poses several challenges. Firstly, the volume of data is huge since graphs
contain up to $10^7$ vertices and $3 \cdot 10^8$ edges (see Tables~\ref{tab:sizes_nodes} and
\ref{tab:sizes_edges} in the Supplementary material for details). Secondly, degree distributions in the Internet tend to follow
the \emph{power law}\cite{Adamic2000, Siganos2003, Faloutsos2003}, which means that there are few
vertices with a large number of neighbors. As a result, even a local neighborhood of distance one can
have a large number of vertices. Thirdly, it is known that malicious activities are localized in the
graph~\cite{Yu2010,collins2007using}, forming communities with sharp boundaries. This implies that
any successful method for detection of malicious activities should be able to aggregate mainly local
information in the graph. In the example in~Figure~\ref{fig:threatcrowd} we observe that the two
second-level domains with repeated characters (a common strategy to trick users with small screens) engage with a
similar set of vertices in the graph.

In the next section, we describe relevant prior art for this problem in the cybersecurity domain.

\section{Related work}%
\label{sec:related_work}


\subsection{Graphical probabilistic models}%
\label{sub:graphical_probabilistic_models}
Graphical probabilistic models capture joint probability distribution of random variables, where
each random variable corresponds to a vertex in a graph, and an edge between two vertices signifies a
dependence between the corresponding variables. High modeling expressivity comes with a price of
exact inference being tractable only for specific subclasses of graphs~\cite{Baum1966, Pearl1988},
therefore for practical problems one has to resort to approximate methods~\cite{Frey, Wang2000a,
Kohli2005}. The most popular inference algorithm is (loopy) belief
propagation~\cite{Kschischang2001}, which iteratively updates states of individual vertices using
states of their neighbors from the previous iteration. This approach is an 
essential component in most following methods~\cite{Carter2013, Carter2014a,Hamilton2017,Chang2015,
Gui2016, Fu2017}. The general message-passing update can be formulated as:
%
\begin{equation}
    \label{eq:message_passing}
    \mathbf{h}_{t}(v)=r\left(\!\widehat{f}\left(\mathbf{h}_{t-1}(v)\right),
        a\left(\!\left\{\!\widetilde{f}\left(\mathbf{h}_{t-1}(u)\right) \mid \forall u \in
    \mathcal{N}(v)\right\}\!\right)\!\!\right)\,,
\end{equation}
where $\widetilde{f}$ first projects representations $\mathbf{h}_{t-1}$ of vertices $u \in \mathcal{N}(v)$ (neighborhood of vertex $v$), function $a$ then aggregates the results, function $\widehat{f}$ projects representation $\mathbf{h}_{t-1}$ of vertex $v,$ and finally, function $r$ operates on a projection of the vertex $v$ together with the output of the aggregation to create a new representation $\mathbf{h}_t.$ Concrete algorithms differ in the definition of $ \widehat{f} $, $
\widetilde{f}$, $a$, and $r.$ 

\subsection{Graph Neural Networks}%
\label{sub:graph_neural_networks}
Graph Neural Networks is a family of methods designed for processing graphs with features on
vertices or edges. We briefly mention two complementary approaches: Spectral and Spatial, and refer the
reader to recent overviews~\cite{Zhou2018a, Wu2019} of this rapidly expanding field. 

\emph{Spectral}~\cite{Bronstein2016,Kipf2016} methods embed a graph into the Euclidean space using the
graph convolution operator. They can reflect structural dependencies among vertices and their features, but they are restricted to the topologies they were trained on. They are mainly
used for reasoning over the whole graph rather than over individual vertices.

\emph{Spatial} approaches leverage spatiality in the graph and define graph
convolution based on vertex relations. GraphSAGE~\cite{Hamilton2017}, one of the first
representatives of spatial methods, defines graph convolution as a refined message-passing update
using neural networks, running for $ T $ iterations:
\begin{equation}
    \label{eq:graph_sage}
    \mathbf{h}_{t}(v) =\sigma\left(\!\mathbf{W}_{t}
    \left[\mathbf{h}_{t-1}(v),a_{t}\left(\!\Big\{\mathbf{h}_{t-1}(u) \mid \forall u \in
\mathcal{N}(v)\Big\}\!\right)\!\right]\!\right)\,,
\end{equation}
where $ a_t $ is a parametrized permutation-invariant aggregation function, and $ \mathbf{h}_0(v) $
is initialized to input features  $ \mathbf{x}_v $ for each vertex. Note the similarity of the update
function~\eqref{eq:graph_sage} to that of the general message passing
algorithm~\eqref{eq:message_passing}. Hence, the spatial GNN can be seen as a general approach to
learning the update function.

GNNs were recently used to solve problems in the cybersecurity domain, for example in\cite{Busch2021, Herath2022, Wei2022, Zhu2022}. \emph{Heterogeneous} GNNs operate on heterogeneous graphs with various vertex and edge types. To deal
with the heterogeneity, many different methods were proposed~\cite{Chang2015, Gui2016, Fu2017,
Shi2017, Dong2017, Zhang2018, Sun2018, Fan2019, Wang2019}, all of which are based on the message-passing paradigm.

\subsection{Probabilistic Threat Propagation}%
\label{sub:probabilistic_threat_propagation}
Probabilistic Threat Propagation (PTP)~\cite{Carter2013, Carter2014a} is a classical graph algorithm
from cybersecurity designed for modeling interactions of vertices in the Internet. Given a set of
known malicious vertices (denylist) it estimates the \emph{threat}, $P(v),$ of each other vertex $v$ in the
graph based on their connections to other vertices. The algorithm defines $P(v)$ as a solution to a
set of linear equations:
\begin{equation}
\label{eq:PTP1}
    P(v_i) = \sum\limits_{v_j \in \mathcal{N}(v_i)} w_{ij} P(v_j | v_i = 0)\,,
\end{equation}
where $w_{ij}$ are positive edge weights assumed to be normalized to one, $\sum_j w_{ij} = 1$.
Conditioning $v_i$ to be benign avoids unwanted direct feedback of a vertex to itself. Since
exact solving of~\eqref{eq:PTP1} requires $ O(n^2) $ time for a graph with $n$ vertices, the authors
propose a message-passing algorithm to find an approximate solution. It initiates $P_0(v)$ to $ 1 $
for vertex from the denylist and $ 0 $ otherwise, and updates the solution in $ T $ iterations as
\begin{equation}
\label{eq:PTP2}
    P_t(v_i) = \sum\limits_{v_j \in \mathcal{N}(v_i)} w_{ij} \left(P_{t-1}(v_j) - C_{t-1}(v_i, v_j)\right)\,,
\end{equation}
where $C_{t-1}(v_i, v_j)$ is a portion of $P_{t-1}(v_j)$ propagated from vertex $v_i$ in the previous
step. In each iteration, $P_t(v)$ on denylisted vertex is set to $ 1 $ again to reinforce the signal.
This is a special case of the message passing procedure in Equation~\eqref{eq:message_passing}.

PTP was successfully used to infer malicious second-level domains from a bipartite DNS graph. The most scaled and
comprehensive study of PTP is in~\cite{Jusko2017}, where it was used for malicious domain discovery. Unlike most of the art,  \cite{Jusko2017} studies how the construction of
graph and setting of edge weights affects the accuracy of identification of malicious domains. We, therefore, consider~\cite{Jusko2017} to be state of the art for malicious domain detection and compare it with the proposed method in Section~\ref{sec:experiments}.

\subsection{Drawbacks for Internet mapping}
While the previous approaches achieved success in their domains, they are not sufficiently flexible to model numerous interactions in the Internet.

The main drawback of PTP is that it can propagate only the single scalar variable $\mathbf{h}_t(v)=P(v)$
and edges can be characterized only by a scalar weight, which limits the description of entities and
their interaction.  Moreover, PTP does not use any statistical learning to optimize the setting of
the algorithm for a given problem and domain leading to potentially inferior
performance~\cite{Jusko2017}.

Methods based on GNNs do not suffer from the limitations of PTP as they allow vertices and edges to be
characterized by feature vectors. However, a large amount of computation is needed to obtain the
result for a single vertex. Because states $ \mathbf{h}_t(v) $ of any vertex $ v $ at any time $ t $
are always the same, the most suitable way to run GNNs is to synchronously compute the results for
all vertices at once, enabling effective state reuse. For the type of data encountered in the
cybersec domain (the dataset used in this work contains tens of millions of vertices), this makes the
inference expensive and training very expensive, especially if results are needed only for a subset
of all vertices. The frequent need for retraining only exacerbates this. GNNs have been used on
large-scale problems~\cite{Rossi2020}, however, speedups were achieved at the expense of 
expressivity.

To tackle the above problems, the proposed method has the following properties:
\begin{itemize}
    \item \emph{Local Inference} --- the inference is always computed w.r.t. to one individual
        vertex, which we call a \emph{central vertex} of the inference, taking its local neighborhood as input (this decreases the complexity of the training).
    \item \emph{Streamlined neighborhood subgraph} --- any      edge from the neighborhood of a central vertex 
        that does not lie on any shortest path to any vertex in the neighborhood is discarded. The
        resulting streamlined neighborhood subgraph is processed using the HMILnet architecture~\cite{Mandlik2021}.
    \item \emph{State heterogeneity} --- the model used for every central vertex is the same (uses the same parameters), however, the internal vertex representations (i.e. the states
        $\mathbf{h}_t(v)$) are \emph{not} shared, are conditioned on the central vertex, and are thus evaluated for each central vertex individually.
\end{itemize}

The above-mentioned assumptions allow us to scale the approach to large heterogenous graphs, which
are ubiquitous in computer security and the Internet in general, or even infinite graphs (e.g.
streamed data).

\section{HMILnet-based graph inference}%
\label{sec:hmilnet_based_graph_inference}
This section briefly reviews the Multiple Instance Learning (MIL) problem and its hierarchical
extension \emph{HMIL}. Then it introduces the \emph{HMILnet} architecture for solving HMIL problems
and discusses specifics of HMIL use for homogeneous graph inference, the concepts of \emph{state
heterogeneity} and \emph{local inference}, and differences to standard GNNs.

\subsection{Multiple Instance Learning}%
\label{sub:multi_instance_learning}
Multiple Instance Learning (MIL)~\cite{Dietterich1997,Maron1998} assumes that each sample called
\emph{bag}, $b$, consists of an arbitrarily large unordered set of vectors $ \{ \mathbf{x}_i
\}_{i=1}^{\vert b \vert} $ called \emph{instances}. Even though it can be assumed that each instance
$ \mathbf{x}_i $ is labeled, these labels are not observed, and the main goal is to infer a label
only for the whole bag $ b $. Many different approaches to solving MIL problems have been proposed,
and we refer an interested reader to reviews in~\cite{Amores2013, Foulds2010}.
\begin{figure}
    \centering
    \includegraphics[width=\columnwidth]{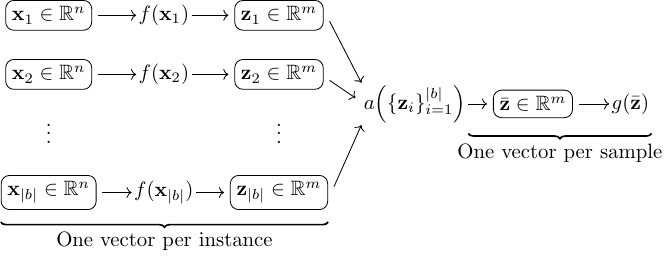}
    \caption{A depiction of a model to solve MIL problems used in~\cite{Pevny2017a, Edwards2016}.}\label{fig:mil-outline}
\end{figure}
Authors of~\cite{Pevny2016,Pevny2017a,Edwards2016} have independently proposed an approach for solving MIL problems, which uses two multi-layer perceptrons, $f$, $g$, with an element-wise
aggregation function $a$ sandwiched in between. Specifically, the whole model is defined as
follows:
\begin{equation}
    g\left(\!a\left(\!\left\{f(\mathbf{x}_i) \mid
    \mathbf{x}_i \in b\right\}\!\right)\!\right)\,.
    \label{eq:bag_model}
\end{equation}
See Figure~\ref{fig:mil-outline} for a more detailed illustration. If aggregation function
$a$ is differentiable, the whole model is differentiable and therefore optimizable by gradient-based
methods. Moreover, \cite{Pevny2019} generalizes the universal approximation theorem for
standard feedforward neural networks to the proposed MIL model.

The two simplest aggregation functions are element-wise $\mean$ and $\max$. Although
\cite{Pevny2019} proves that $\mean$ aggregation is theoretically sufficient in the MIL framework,
\cite{Pevny2017a} discusses that in some situations $\max$ can perform better.

\subsection{Hierarchical Multiple Instance Learning}%
Hierarchical Multiple Instance Learning (HMIL) is an intuitive, non-compromising and theoretically
justified~\cite{Pevny2019} framework useful for processing hierarchically structured data.

\label{sub:hierarchical_mil}
MIL problem is extended in~\cite{Pevny2019} to hierarchical cases, where instances in
bags can be bags themselves or Cartesian products of such spaces. Specifically, the class of
HMIL samples can be defined recursively as follows:
\begin{itemize}
    \item\emph{Leafs:} Feature vector $ \mathbf{x} \in \mathbb{R}^n$ is an HMIL sample.
    \item\emph{Bags:} If instances $x_1, \ldots, x_k$ are HMIL samples following the same
        \emph{schema} (e.g.~all are vectors of the same length or bags of vectors of the same
        length), then bag $ \{ x_i \}_{i=1}^{k} $ is also an HMIL sample.
    \item\emph{Tuples:} If $x_1, \ldots, x_l$ are any HMIL samples, then $(x_1, \ldots, x_l)$ is also an HMIL sample.
\end{itemize}
The extension is important because the class of HMIL samples includes data omnipresent in Internet
communication and data exchange such as XMLs or JSONs.

Note that whereas instances in bags are required to be structured identically (follow the same
schema), this is not required for tuples.


For a set of HMIL samples following the same schema, a corresponding HMIL model
is built by combining three types of layers:
\begin{itemize}
    \item\emph{Leaf layer} $ f $: a mapping that accepts a vector $ \mathbf{x} $ from a vector
        space $\mathbb{R}^n $ and maps it to $\mathbb{R}^o$.
    \item\emph{Bag layer} $ (a, g) $: accepts a set of vectors $b = \{ \mathbf{z}_i
        \}_{i=1}^{k}$ of the same length $ m $ and processes them as $g(a(b))$, where $ a $ is an
        element-wise aggregation function and $g \colon \mathbb{R}^m \mapsto \mathbb{R}^o$ a
        mapping.
    \item\emph{Product layer} $ r $: accepts an $ l $-tuple of arbitrary vectors $\mathbf{z}_1, \ldots,
        \mathbf{z}_l$, where $ \mathbf{z}_i \in \mathbb{R}^{m_i} $, and processes them as $r\left(\left[
        \mathbf{z}_1,\ldots, \mathbf{z}_l \right] \right)$, where $ r\colon \mathbb{R}^{\sum_i m_i}
        \mapsto \mathbb{R}^o $. Here, $[\ldots]$ denotes concatenation.
\end{itemize}
The composition is enabled by all three layer types outputting fixed-size vectors.

Bag with instances of an arbitrary (but same) structure are projected to a vector by first
projecting each instance with the \emph{same} HMIL (sub)model consisting of appropriate layers to
a vector of the same length, and then applying an aggregation function (e.g. $\mean$ or $\max$) to obtain one vector representing the whole bag. Equation~\eqref{eq:bag_model} is an example of a model where a leaf
layer $f$ is followed by a bag layer formed from $(a, g)$.

A heterogeneous tuple with items of \emph{different} structure is processed by first projecting each item to a vector with a \emph{different} HMIL (sub)model and then concatenating
the results to create a single vector.

The complete model, tailored specifically for each hierarchy of data, is differentiable with respect
to its parameters if all its components are. Therefore in application it is recommended to
implement $ f $,  $ g $, and  $ r $ as one or more layers of fully connected neural network layers.
Such architecture is called~\emph{HMILnet}.

Many applications have missing data, which in the case of HMIL means empty bags,  missing items in
tuples, and missing values in leaves. In all these cases, empty parts are imputed with learnable
parameters, which values are optimized during training in the same way as other model parameters.

HMILnet can be viewed as a feature extractor optimized for given hierarchical data (in the sense
of the above definition). Examples of successful applications of the HMIL paradigm in computer security
can be found in~\cite{Pevny2016, Pevny2020, Mandlik2021, Mandlik2021a}.

\subsection{Adapting HMILnet to graph inference}%
\label{sub:hmilnet_for_graph_inference}
The idea behind the use of HMIL for graph inference is to interpret neighborhood of a fixed
(\emph{central}) vertex (provided as \emph{input} to the algorithm) as a \emph{bag}. Both instances in bags and vertices in the neighborhood of
a central vertex are unordered and can be arbitrarily large, which justifies the use of HMILnet to
learn values about the central vertex~\cite{Pevny2017a}.

We now describe how the HMIL paradigm is used in the processing of homogeneous graphs, and the extension to heterogeneous graphs is deferred to the next section.  After defining several useful terms, we describe HMIL for a neighborhood of a distance one as the simplest case and then proceed to the general
definition.

We assume an \emph{undirected homogeneous graph} $ G = (V,  E) $ containing a set of \emph{vertices} $V$ and a set of \emph{edges} $E$, where $E \subseteq {V \choose 2}$. We define each edge $\{u, v\} \in E$ as an \emph{unordered set} of its two endpoints, vertices $u, v \in V$. Moreover, we assume that each vertex $v$ and each edge $\{u, v\}$ is attributed feature vectors $ \mathbf{x}_v $, $ \mathbf{x}_{v, u} $, respectively. A single-step \emph{neighborhood} of a single vertex $ v \in V$ is defined as
\begin{equation}
    \mathcal{N}(v) = \mathcal{N}_1(v) = \Big\{u \in V \mid \left\{u, v\right\} \in E
    \Big\}\,.
\end{equation}
A \emph{$ t $-step neighborhood} $ \mathcal{N}_t(v),$ $ t > 1 $, is a generalization defined as the set of
vertices whose distance from $ v $ is \emph{exactly}  $ t $. Finally, the set of vertices with a distance at most $t$ is denoted as
$\mathcal{N}_{\leq t}(v) = \bigcup_{i = 0}^t \mathcal{N}_i(v).$
\emph{Streamlined $t$-step neighborhood subgraph} of a vertex $v$ contains vertices $
\mathcal{N}_{\leq t}(v) $ and only edges lying on one of the shortest paths from $ v $ to any vertex in  $ \mathcal{N}_{\leq t}(v) $.

The inference is then introduced for a vertex $v$ (called \emph{central vertex}), which is considered fixed. HMIL sample for a single-step neighborhood of a \emph{central vertex} $ v \in V $ is defined as a tuple $(v, b(v)),$ where $b(v)$ is a bag containing information from vertices and edges in the neighborhood defined as 
\begin{equation}
    b(v) = \Big\{\mathbf{z} = [\mathbf{x}_u, \mathbf{x}_{v, u}] \mid u \in \mathcal{N}(v)\Big\}\,,
    \label{eq:bag_graph}
\end{equation}
where $[\cdot, \cdot]$ denotes concatenation. The corresponding HMILnet model is defined as 
\begin{align}
    \mathbf{b}(v) &= g\left(\!a\left(\!\left\{\widetilde{f}(\mathbf{z}) \mid \mathbf{z} \in
    b(v)\!\right\}\!\right)\!\right) \label{eq:bagrep} \\
    \mathbf{h}(v) &= r\left(\!\left[\widehat{f}(\mathbf{x}_v), \mathbf{b}(v)\right]\!\right)\,,
    \label{al:step1}
\end{align}
consisting of two leaf layers $\widehat{f}$ and $\widetilde{f}$, a bag layer $(a, g),$ and a product layer $r.$ 

The model first processes each vertex $u \in \mathcal{N}(v)$ (an instance in bag $ b(v) $) with a
leaf layer $ \widetilde{f}$, and then by applying a bag layer $ (a, g) $ we obtain a representation of the bag $
\mathbf{b}(v)$. In parallel, a representation of the central vertex $ v $ is obtained by applying a
leaf layer  $ \widehat{f} $ on its feature vector $\mathbf{x}_v$. Finally, we concatenate the two outputs together using a product layer $ r $.

When extending the above for processing two-step neighborhoods,
the single-step construction is repeated twice, but importantly over the streamlined neighborhood
subgraph of the central vertex $v.$ For each vertex $u \in \mathcal{N}_1(v),$ HMIL sample $(u,
b(u))$ is then defined as described above. Since the construction is in the streamlined neighborhood
subgraph, the bags $ b(u) $ contain vertices that neighbor $u$ and are at the distance of two from $v$ as well as feature vectors corresponding to edges leading to these vertices. Finally, another HMIL sample is produced for a central
vertex $v$, where feature vectors $\mathbf{x}_u$ of vertices $u \in \mathcal{N}(v)$ are
replaced with HMIL structures $b(u)$ created in the previous step. The HMILnet model performing inference
is constructed accordingly.

The general algorithm for $ T $-step HMILnet-based inference on a graph is described in
Algorithm~\ref{alg:inference}. Note that no leaf layers $ \widetilde{f}_t $ and $ \widehat{f}_t $,
bag layers $ (a_t, g_t) $, or product layers $ r_t $, in the HMILnet inference model share
parameters.

\begin{algorithm}
    \caption{$ T $-step HMILnet-based inference on a (homogeneous) graph}%
    \label{alg:inference}
    \begin{algorithmic}[1]
        \renewcommand{\algorithmicrequire}{\textbf{Input:}}
        \renewcommand{\algorithmicensure}{\textbf{Output:}}
        \REQUIRE $ T $, graph $ G = (V, E) $, vertex features $ \mathbf{x}_u $, edge features  $
        \mathbf{x}_{u, w} $, central vertex  $ v \in V $, HMILnet model with layers $
        \widetilde{f}_t, \widehat{f}_t, (a_t, g_t), r_t$ for each $ t = 1, \ldots, T$
        \ENSURE $ \mathbf{h}_T(v) $
        \STATE $ \forall u \in V\colon \mathbf{h}_0(u) = \mathbf{x}_u $
        \FOR {$t = 1$ to $T$}
        \FORALL {$u \in \mathcal{N}_{T-t}(v) $}
        \STATE $ b_t(u) = \Big\{\!\left[\mathbf{h}_{t-1}(w), \mathbf{x}_{u, w}\right]\!\mid\! w \in \mathcal{N}(u)
        \cap\mathcal{N}_{T-t+1}(v)\!\Big\} $

        \STATE $ \mathbf{b}_t(u) = g_t\left(\!a_t\left(\!\left\{\widetilde{f}_t(\mathbf z) \mid \mathbf z \in
        b_t(u)\right\}\!\right)\!\right) $
        \STATE $ \mathbf{h}_t(u) = r_t\left(\!\left[\widehat{f}_{t}(\mathbf{x}_v),
        \mathbf{b}_t(u)\right]\!\right) $
        \ENDFOR
        \ENDFOR
        \RETURN $\mathbf{h}_T(v)$
    \end{algorithmic} 
\end{algorithm}

The training of the model follows the standard machine-learning procedure --- collecting one or
more graphs, extracting some relevant central vertices and their neighborhoods to form a minibatch,
deriving labels from the denylist, and updating the whole model w.r.t. some loss.

\subsection{Differences between HMILnet and GNN}%
\label{sub:gnn_comparison}
The first distinction between modes of operation of GNNs and HMILnet-based inference is the \emph{state
heterogeneity}. GNNs enforce the property that for any vertex $ v $ at any time $ t $, $
\mathbf{h}_t(v) $ is always the same, whereas in HMILnet-based inference  $ \mathbf{h}_t(v) $ depends on the current central vertex. This is a consequence of neighborhood streamlining.

Therefore, if enough memory is available, the most efficient way to evaluate GNN is \emph{global
inference}---hold all states $ \mathbf{h}_t(v) $ in memory, synchronously compute all states $
\mathbf{h}_{t+1}(v) $ at once, and, having done all $ T$ steps, output final representations $
\mathbf{h}_T(v) $ for all vertices in the graph.

Contrary, HMILnet-based inference uses the \emph{local inference} computational model, where the
result $\mathbf{h}_T(v) $ is computed separately for each vertex $v.$ The local inference is less
computationally efficient when one is interested in inference of a large number of vertices since the
computation is not reused. But when one is interested in a small fraction of vertices, as is the case
of our application,  local inference is more efficient.

Whereas GNNs draw most inspiration from convolutional networks with each layer estimating graph
convolution operation, which leads to powerful but hard-to-scale models, the HMILnet-based approach
regards the graph (or a collection of graphs) merely as a \emph{database} for querying information
about vertices and their relationships. This is beneficial in problems with large graphs and high
degrees of vertices, and in particular in
computer security by the fact that the maliciousness of the vertex is mainly influenced by its local
neighbors with little dependence on the global graph structure.

Because local inference in the HMILnet-based algorithm treats each central vertex as a single training
`example', it is more straightforward to construct minibatches for training models and to employ all
additional sampling techniques for vertices like stratified sampling or prioritized sampling.

\section{Mapping the Internet}%
\label{sec:mapping_the_internet}
This section describes concrete application of the above framework to the problem of detection of malicious second-level domains from their interactions in the Internet. However, the proposed algorithm is \emph{general} and it could be employed for processing any other type of network objects. For brevity from now on, we always refer to `second-level domains' as simply `domains'. 

We first describe the input and then the construction of \emph{transformed graphs}, which are then used to construct HMIL samples and the corresponding HMILnet model. The remaining part of the section is devoted to implementation aspects.

\subsection{Input}%
\label{sub:input}
The method assumes as input an undirected heterogeneous graph similar to the one in
Figure~\ref{fig:threatcrowd}, with several vertex types corresponding to different network entities
like domains, network clients, IP addresses, executables, emails, and others. The interest is on
domains, denoting a set of all observed domains $ \mathcal{D} $, for which a denylist of known
malicious domains $ L \subset \mathcal{D} $ is available (considering non-denylisted domains
`unknown'). The task is to predict whether the remaining domains from $ \mathcal{D} \setminus L $
are \texttt{malicious} or \texttt{benign}.

Vertices of the heterogeneous graph(s) can be partitioned into sets $ \mathcal{D},
\mathcal{V}^{(1)}, \ldots, \mathcal{V}^{(C)} $, where $\mathcal{D}$ is a set of vertices
corresponding to domains and  $\mathcal{V}^{(i)}$ are sets of vertices corresponding to $C$ network
entity types. The transformed graphs described below are constructed from subgraphs $(\mathcal{G}^{(1)},
\ldots, \mathcal{G}^{(C)})$ corresponding to the interaction of domains with one type of network entity.
Each subgraph $\mathcal{G}^{(i)} = ( \mathcal{D} \cup \mathcal{V}^{(i)}, \mathcal{E}^{(i)})$, is
\emph{bipartite}---for each edge $\{d, v\} \in \mathcal{E}^{(i)}$ it holds $d \in \mathcal{D}$ and
$v \in \mathcal{V}^{(i)}$. 

For instance, one specific $ \mathcal{V}^{(i)} $ can be a set of vertices representing network clients
with edge $ \{ d, v \} \in \mathcal{E}^{(i)}$ meaning that client $v \in \mathcal{V}^{(i)}$
communicated with domain $d \in \mathcal{D}$ during a predefined time window in which the network
was observed~\cite{Carter2014a, Jusko2017}.

In different words, each graph $ \mathcal{G}^{(i)} $ also represents a \emph{homogeneous binary relation} between entities from $ \mathcal{D} $ and $ \mathcal{V}^{(i)} $.

\subsection{Transformed graph}%
\begin{figure}
    \centering
    \includegraphics[width=\columnwidth]{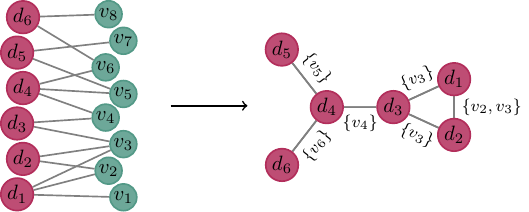}
    \caption{Graph transformation of a bipartite graph $ \mathcal{G}^{(i)} $ (on the left) to a \emph{transformed graph}
    $ G^{(i)} $ (on the right). Edges in the transformed graph are labeled with the names of vertices that
    interacted with both incident vertices in the original graph. Note that vertices $v_1$, $v_7$, and
    $v_8$ all have only one neighbor and thus do not influence the resulting transformed graph.}%
    \label{fig:transformation}
\end{figure}

\label{sub:transformed_graph}
To decrease the computational complexity of the inference (and training), graphs $\mathcal{G}^{(i)} = (
\mathcal{D} \cup \mathcal{V}^{(i)}, \mathcal{E}^{(i)})$ are converted to \emph{transformed graphs}
$G^{(i)} = ( \mathcal{D}, E^{(i)} )$ with only domains $\mathcal{D}$ as vertices. Vertices corresponding
to other network entities are discarded, and edges in the transformed graph exist iff
\begin{equation}
    \label{eq:transformation}
    \left\{ d_1, d_2 \right\} \in E^{(i)} \iff \exists v \in \mathcal{V}^{(i)} \colon \left\{\left\{d_1, v\right\}, \left\{d_2, v\right\}\right\}
    \subseteq \mathcal{E}^{(i)} ,
\end{equation}
for all different $d_1, d_2 \in \mathcal{D}$. In other words, we put edge connecting $d_1$ and $d_2$ into a \emph{transformed} graph
if and only if there exists a vertex $v$ that is connected to both in the \emph{original} graph. Figure~\ref{fig:transformation}
illustrates the process.

For example, if $ \mathcal{V}^{(i)} $ are network clients, the transformed graph contains an edge
between two different domains if and only if there exists a client that communicated with both of
them. Therefore, each client creates a clique in the transformed graph containing each contacted
domain. For instance, in Figure~\ref{fig:threatcrowd}, the big gray vertex representing IP address
\texttt{23.229.206.69} will connect all eight red vertices representing domains like
\texttt{technewsup.com} and \texttt{beyondgraphicsllc.com} into one clique in the transformed graph
of domains.

As a result, only domains that have at least one interaction in common with another entity in the network, are connected in the transformed graph. Further motivation for this transformation can be found in~\cite{Liu2014, Jusko2017, Manadhata2014}.

\subsection{Inference}%
\label{sub:inference}
Let's now assume a fixed central vertex $d \in \mathcal{D}$, whose neighborhood we want to describe
by an HMIL sample, such that the inference can be efficiently performed by an HMILnet model. The HMIL
sample is created by first creating the streamlined $ T $-step neighborhood subgraph of $d$ in each of the $
C $ transformed graphs, from which $C$ HMIL (sub)samples $ x_1, \ldots, x_C $ are extracted as
described in Section~\ref{sub:hmilnet_for_graph_inference}. Thus, each HMIL subsample $ x_i $
describes relations of central vertex $d$ with one type of entities in its $T$-step neighborhood. The
final HMIL sample is then constructed as a tuple of HMIL subsamples $ (x_1, \ldots, x_C) $. Finally,
an HMILnet model for inference is created to match the sample as described in
Section~\ref{sec:hmilnet_based_graph_inference} (using a product layer stacked on top for processing
the tuple). Since the goal is to classify domains into \texttt{malicious} and \texttt{benign} classes,
the final layer is of dimension two, corresponding to the predictive probabilities.

\begin{figure*}
\begin{center}
\includegraphics[width=\textwidth]{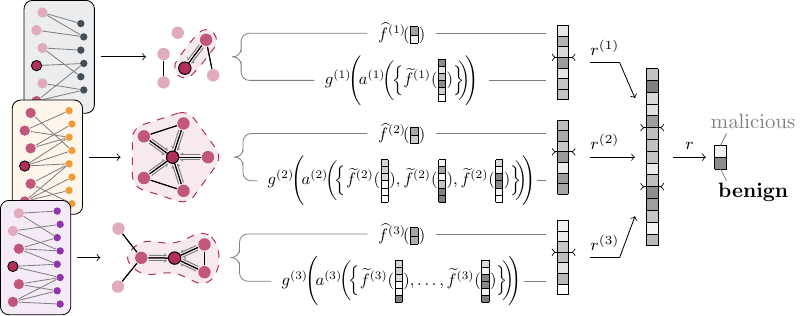}
\caption{
    The whole model procedure for the case when $ T = 1 $. Input bipartite graphs $ \mathcal{G}^{(i)} $
    representing input binary relations on the left are first used to obtain the same number
    of transformed graphs $ G^{(i)} $. The central vertex representing one particular domain is
    highlighted in each graph as well as its neighbors in the transformed graph. Note that the left
    part of each bipartite graph and each transformed graph consist of an identical set of vertices (in
    our case representing domains) as opposed to edges, which differ with each
    relation. In the next phase, we aggregate vertex and edge features as explained in
    Section~\ref{sec:hmilnet_based_graph_inference}. Each of the three graphs uses (sub)models $
    \widehat{f}^{(i)}, g^{(i)}, a^{(i)}, \widetilde{f}^{(i)}, r^{(i)} $ that do not share parameters and may
    even have different topology. Finally, the product construction is done to obtain the final
    output, which in the 1-step case is interpreted as a vector of predictive probabilities. Best
    viewed in color.
}%
\label{fig:full_model}
\end{center}
\end{figure*}

A sketch of the whole procedure for one step-neighborhood ($T = 1$) is shown in
Figure~\ref{fig:full_model}. The pseudocode of the inference is in Algorithm~\ref{alg:full}.
\begin{algorithm}
    \caption{ Heterogeneous mapping of the Internet }%
    \label{alg:full}
    \begin{algorithmic}[1]
        \renewcommand{\algorithmicrequire}{\textbf{Input:}}
        \renewcommand{\algorithmicensure}{\textbf{Output:}}
        \REQUIRE central vertex (domain) $ d $, heterogeneous graphs $(\mathcal{D} \cup\mathcal{V}^{(1)}, \mathcal{E}^{(1)}), \ldots,(\mathcal{D} \cup\mathcal{V}^{(C)}, \mathcal{E}^{(C)})$, HMILnet models for graph inference $ M^{(1)}, \ldots,
        M^{(C)}$, number of inference steps $ T $, final product layer $ r $
        \ENSURE binary classification logits $ \mathbf{o} $
        \FOR[compute transformed graphs] {$i = 1$ to $C$}
        \STATE $\begin{aligned}
            E^{(i)} = \big\{\left\{ d_1, d_2 \right\} \mid & \left\{d_1, v\right\},
            \left\{d_2, v\right\} \in \mathcal{E}^{(i)}, \\ & d_1, d_2 \in \mathcal{D}, v \in \mathcal{V}^{(i)}\big\}
        \end{aligned}$
        \STATE $ G^{(i)}= (\mathcal{D}, E^{(i)}) $
        \ENDFOR
        \FOR[run HMILnet-based inference] {$i = 1$ to $C$}
        \STATE $\mathbf{h}_T^{(i)} = \text{graph-inference}(d, G^{(i)}, M^{(i)}, T)$
        \ENDFOR
        \STATE $ \mathbf{o} = r\left(\!\left[\mathbf{h}^{(1)}_T, \ldots, \mathbf{h}^{(C)}_T\right]\!\right) $ \COMMENT{aggregate}
        \RETURN $\mathbf{o}$
    \end{algorithmic} 
\end{algorithm}

The described procedure infers a value for a \emph{single} central vertex (single training example), which we regarded as \emph{input} to the Algorithm~\ref{alg:full}. In the experiments presented in this work, the central vertex corresponds to a single domain. Minibatches during training contains 256 independently sampled samples (each sample corresponds to one central vertex and his neighborhood).

\subsection{Graph features}%
\label{sub:features}
In Section~\ref{sec:hmilnet_based_graph_inference}, it was mentioned that HMILnet model for graph inference can implicitly handle rich features on vertices and edges. Although additional features extracted
from individual entities could be added, we have not pursued this idea and extracted features only
from raw interaction observations encoded in input graphs $ \mathcal{G}^{(i)} $ (or their
transformations $ G^{(i)} $). This is motivated by the fact that in the computer security domain
feature extraction tends to be a complicated and time-consuming process. Overall, this decision
makes the task harder, as the signal from binary relations tends to be noisy and less
discriminative, however, it is less demanding on data collection, processing, and domain knowledge.
Finally, this decision enables fair comparison with the PTP algorithm.

By enriching the graph with features extracted from the graph, HMILnet can select informative features
or their combination during training. This selection is done manually in~\cite{Jusko2017}, where
the author searches for optimal construction of graph for PTP algorithm~\cite{Carter2014a}.
Table~\ref{tab:features} summarizes all vertex-level $ \mathbf{x}_v $ and edge-level $
\mathbf{x}_{v, u} $ features used in experiments below. The features are a superset of features
considered in~\cite{Jusko2017}. Features without an upper bound (e.g. vertex degrees) are mapped by
$ x \mapsto \log(x + 1) $ to reduce their dynamic range.


%
\begin{table}
    \caption{Specification of features used in the experiments. Here, $ \mathcal{G} $ and $ G $
        denote (original) bipartite and transformed graphs, and $\mathcal{N}_{\mathcal{G}}(v)$ and
    $\mathcal{N}_{G}(v)$ neighborhoods of $ v $ in graphs $ \mathcal{G} $ and $ G $.}%
    \label{tab:features}
    \centering
    {\renewcommand\arraystretch{1.2}
        \begin{tabular}{cccl}
            \toprule
            name & part of & definition & range\\
            \midrule
            degree & $\mathbf{x}_{v}$ & $\left\vert \mathcal{N}_{\mathcal{G}}(v) \right\vert$ &
            $[1, \infty)$ \\
            transformed degree & $\mathbf{x}_{v}$ & $\left\vert \mathcal{N}_{G}(v) \right\vert$ & $
            [0, \infty)$\\
            neighbor degree & $\mathbf{x}_{v, u}$ & $\left\vert \mathcal{N}_{\mathcal{G}}(u)
            \right\vert$ & $ [1, \infty) $ \\ intersection & $\mathbf{x}_{v, u}$ & $\left\vert
            \mathcal{N}_{\mathcal{G}}(v) \cap \mathcal{N}_{\mathcal{G}}(u) \right\vert$ & $ [0, \infty) $\\
            union & $\mathbf{x}_{v, u}$ & $\left\vert \mathcal{N}_{\mathcal{G}}(v) \cup
            \mathcal{N}_{\mathcal{G}}(u) \right\vert$ & $ [1, \infty) $\\
            Jaccard index & $\mathbf{x}_{v, u}$ & $\frac{\left\vert \mathcal{N}_{\mathcal{G}}(v) \cap\
            \mathcal{N}_{\mathcal{G}}(u) \right\vert}{\left\vert \mathcal{N}_{\mathcal{G}}(v) \cup\
            \mathcal{N}_{\mathcal{G}}(u) \right\vert}$ & $ [0, 1] $\\
            attachment & $\mathbf{x}_{v, u}$ & $\left\vert \mathcal{N}_{\mathcal{G}}(v)\right\vert
            \left\vert\mathcal{N}_{\mathcal{G}}(u) \right\vert$ & $ [0, \infty) $\\
            detected & $\mathbf{x}_{v, u}$ & $ u \in L $ & $ \{0, 1\} $\\
                \bottomrule
            \end{tabular}
        }
\end{table}

\subsection{Importance sampling}%
\label{sub:importance_sampling}
The power distribution of degrees of vertices found in many graphs representing social networks or the Internet
means that there is a small group of domains with extremely large neighborhoods (order of millions).
If these domains occur in HMIL samples, bags representing them would have a large number of instances
(Equation \eqref{eq:bag_graph}) leading to a high computational complexity of the graph inference
part. A solution is to not include the full neighborhood in the bag $b(v)$, but only a randomly selected
subset, resorting to approximation of the result of aggregation over the bag. Since most domains in
the neighborhood are likely `unknown' (non-denylisted) and uniform random sampling would likely miss
any informative denylisted malicious domains, we have used importance sampling which selects unknown
and known malicious domains with different probabilities.

\emph{Importance sampling}~\cite{owen2013, kahn1953} is used to approximate the output of
aggregation function $ a $ on bag $ \mathbf{b}(v) $ (line 5 in Algorithm~\ref{alg:inference}).
Positive (malicious) and negative domains are sampled to a smaller bag with different probabilities
(explanation of sampling follows shortly below). Then the $\mean$ aggregation function is modified
to its \emph{weighted} variant, where the positive and negative instances are weighted by weights
$w^{+}$ and $w^{-}$, respectively, to ensure the output is an unbiased estimate of the true
value.

The weights $w^{+}$ and $w^{-}$ and sampling probabilities are set as follows. Let $n^+$ denote the
number of positive  domains in $b(v)$ according to the denylist, and $n^-$ the number of
negative (`unknown') domains in $b(v)$, where $n^+ \ll n^-$. To sample a smaller bag $b'(v)$ from
$b(v)$ we include all $n^+$ positive domains and $\min\left\{ n^-, k^- \right\}$ randomly selected
negative domains, where $k^-$ is a parameter.  In the case when $n^- > k^-$, `unknown' domains are
uniformly sampled without repetition, and each of them is assigned weight $w^- = n^-/k^-$. Each
positive domain is assigned weight $ w^+ = 1 $. 

An additional positive side-effect of the subsampling is an introduction of stochasticity into
learning, which is known to act as a regularizer~\cite{srivastava2014a, ioffe2015}.

\section{Experimental settings}%
\label{sec:experiments}

In the rest of the text, we describe experimental results. First, we specify how the data were
obtained, then various details about implementation and parameter values, and finally, the evaluation
protocol. Specific results are then presented in the next Section~\ref{sec:results}.

\subsection{Data}%
\label{sub:data}
Data were kindly provided by \emph{Cisco Systems, Inc.} and consist of several binary relations
(represented as a bipartite graph defined in Section~\ref{sub:input}) between domains and other
network entities. Three following relations were extracted from a subset of anonymized web proxy
(W3C) logs processed by Cisco Cognitive Intelligence\footnote{\url{https://cognitive.cisco.com}},
and Cisco AMP
telemetry\footnote{\url{https://www.cisco.com/c/en/us/products/security/advanced-malware-protection/index.html}}:
\begin{itemize}
    \item \texttt{domain-client} --- a client has communicated with (issued an HTTP(S) request to) a domain.
    \item \texttt{domain-binary} --- a process running a binary file has communicated with a domain.
    \item \texttt{domain-IP} --- a domain hostname was resolved to an IP address using DNS.
\end{itemize}
The data were further enriched with the following three relations extracted from fields in the
latest TLS certificate issued to the domain\footnote{more specifically to one of its possible
hostnames}:
\begin{itemize}
    \item \texttt{domain-TLS issuer} --- a relation connecting domains to the issuer of the TLS
        certificate. The cardinality type is many-to-one and, therefore, the transformed graph contains a fully connected component for each of the issuers.
    \item \texttt{domain-TLS hash} --- a many-to-one relation connecting a domain to the hash of the TLS certificate. In the transformed graph, domains using the same certificate are connected.
    \item \texttt{domain-TLS issue time} --- domain and the time when the validity period of the certificate starts, stored as a \emph{Unix timestamp}.
\end{itemize}
Finally, from publicly available information in \emph{WHOIS}, six more relations were extracted: \texttt{domains} and \texttt{email}, \texttt{nameserver}, \texttt{registrar
name}, \texttt{country}, \texttt{WHOIS id}, and \texttt{WHOIS creation time}.

Relations \texttt{domain-\{client, binary, IP\}} are of the many-to-many cardinality type---one domain
may be connected with an edge to multiple clients/binaries/IPs and vice versa. The remaining
cardinalities are of the many-to-one cardinality type.

Graphs of all eleven relations were constructed from data collected during one week-long window. All
eleven relations were collected every week. A total of twelve weeks from three months in 2019 were
available for experiments --- \texttt{05-23}, \texttt{06-03}, \texttt{06-10}, \texttt{06-17},
\texttt{06-24}, \texttt{06-26}, \texttt{06-27}, \texttt{07-01}, \texttt{07-08}, \texttt{07-15},
\texttt{07-22} and \texttt{07-29}. Further details (specific examples, graph sizes) are available in
Tables~\ref{tab:relations},~\ref{tab:sizes_nodes}, and~\ref{tab:sizes_edges} in the Supplementary
material. An average bipartite graph describing interaction of domains with a single entity contains
around $7\cdot 10^6$ vertices and $3 \cdot 10^8$ edges.

A denylist of malicious domains, $L$, provided by Cisco Cognitive Intelligence, tracked 335 malicious campaigns. Each campaign represents a different threat type, its variant, or the stage of the attack. Most of campaigns belong to the family of command and control (for families like ransomware or trojans), represent content delivery or download link of different kinds of malware, or misuse the web advertising via click frauds or other kinds of malwertising. Before each domain was added to the denylist, it was reviewed and categorized by a human analyst.

For each window, we used a snapshot of the denylist available at that time
reflecting the knowledge at that time. All weeks of data contained approximately $500$ observed
denylisted domains. Additional statistics about the denylists are in Table~\ref{tab:mal_numbers} in
the Supplementary material. 

\begin{table*}[htb]
    \caption[Numbers of neurons in models from the \emph{Modelling Internet communication}
    experiment.]{Numbers of neurons in model architectures $ \mathcal{M}_b $,  $ \mathcal{M}_w $ and
        $ \mathcal{M}_d $. Each number in tuples represents one feedforward layer with the given
        number of neurons. The length of the tuple specifies the number of layers in the submodel.
        For example, the baseline architecture $ \mathcal{M}_b $ implements leaf layer $
        \widehat{f}^{(i)} $ as a sequence of three dense layers with $2$, $10$, and $10$ neurons.}%
    \label{tab:topologies}
    \centering
    {\renewcommand\arraystretch{1.25}
        \begin{tabular}{clllll}
            \toprule
            (sub)model & $ \widehat{f}^{(i)} $ & $ \widetilde{f}^{(i)} $ & $ g ^{(i)} $ & $ r^{(i)}
            $ & $ r $ \\
            \midrule
            $ \mathcal{M}_b $ & $(2, 10, 10)$ & $(6, 30, 30)$ & $ (120, 60, 60) $ & $ (70, 20)$ & $ (220, 100, 2) $ \\
            $ \mathcal{M}_w $ & $(2, 20)$ & $(6, 60)$ & $ (240, 120) $ & $(140, 20)$ & $ (1320, 2) $ \\
            $ \mathcal{M}_d $ & $(2, 5, 5, 5)$ & $(6, 15, 15, 15, 15)$ & $ (60, 60, 30, 30) $ & $
            (35, 20, 20) $ & $ (220, 60, 60, 2) $ \\
            \bottomrule
        \end{tabular}
    }
\end{table*}

\subsection{Implementation details}%
\label{sub:experimental_setup}
Experimental results presented below were obtained using a one-step ($ T=1 $) neighborhood
(Figure~\ref{fig:full_model}). We expect this to be sufficient in practice since malicious
activities tend to be localized~\cite{Yu2010} and a few-step neighborhood will already cover most of
the graph (traversing through vertices of large degrees). We leave the exploration of multi-step
versions for future work.

We have compared three different architecture settings differing in the topology of submodel
networks implementing $\widehat{f}^{(i)} $, $ \widetilde{f}^{(i)} $, $ g^{(i)} $, and $r^{(i)} $
(see Equation~\eqref{eq:bagrep} and~\eqref{al:step1}). These four submodel networks have the same
topology for each of the eleven input graphs but do not share parameters. Wider $ \mathcal{M}_w $
networks have twice many neurons as the baseline, $ \mathcal{M}_b $, but keep the number of layers the
same. Deeper $ \mathcal{M}_d $ networks have more layers than baseline $\mathcal{M}_b,$ but a similar
number of neurons. Details can be found in Table~\ref{tab:topologies}. All networks used $\tanh$ in
layers just before aggregations $ a^{(i)} $ and $\relu$ otherwise.

Aggregation layers $ a^{(i)} $ implemented concatenation of four element-wise
functions: weighted $ \mean $, $ \max $, parametric \emph{LogSumExp}~\cite{Kraus2015} (a smooth approximation of $ \max $):
\begin{equation}
    \label{eq:lse}
    a_{\lse}(\{x_i\}_{i=1}^k; \theta_r) = \frac{1}{\theta_r}\log \left(\frac{1}{k} \sum_{i = 1}^{k}
    \exp({\theta_r\cdot x_i})\right)\,,
\end{equation}
and parametric weighted $ \pnorm $~\cite{Gulcehre2014}:
\begin{equation}
    \label{eq:pnorm}
    a_{\pnorm}(\{x_i, w_i\}_{i=1}^k; \theta_p, \theta_c) = \left(\frac{\sum_{i = 1}^{k}
    w_i\cdot\vert x_i - \theta_c \vert ^ {\theta_p} }{\sum_{i=1}^k
w_i}\right)^{\frac{1}{\theta_p}}\,,
\end{equation}
where $  \theta_r $, $ \theta_p $, and  $ \theta_c $ are trainable parameters. The output of the
aggregation has four times higher dimension than the input (see Table~\ref{tab:topologies}).

Because some of the bags may be very large, we employed the sampling procedure described in
Section~\ref{sub:importance_sampling} with $ k^- = 100 $. Weighted $\mean$ and $\pnorm$ use the
resulting importance sampling weights $ w_i $ in the computation. If for a domain some relations are
not available at all, or the domain has no neighbors in one or more transformed graphs, the
neighborhood reflects in an empty bag treated as described in Section~\ref{sub:hierarchical_mil}.

All parameters $\bm{\theta}$ of the model therefore are (i) weights and biases used in all feedforward
layers, (ii) vectors of default values for cases of missing data or empty neighborhoods in the
transformed graph, and (iii) parameter vectors ($ \bm{\theta}_r $, $ \bm{\theta}_p $ and $
\bm{\theta}_c $) used in aggregation mappings.

Minibatches contained $256$ central vertices sampled without repetition containing an equal number of
positive and negative examples. One thousand minibatches were created from each
graph in one epoch. The training ran for five epochs. Since low false positive rate is paramount
in computer security, the loss function was weighted binary cross entropy
\begin{equation}
\begin{aligned}
    \label{eq:weighted_crs_entropy}
    \mathcal{L}(\bm{\theta}) = - \frac{1}{n} \sum_{i=1}^{n} \Big(&\omega_1 y_i \log p(\widehat{y}_i = 1; \bm{\theta})
    + \\ &\omega_0(1-y_i) \log p(\widehat{y}_i = 0; \bm{\theta})\Big),
\end{aligned}
\end{equation}
where $n$ is the size of the minibatch, $ y_i \in \left\{ 0, 1 \right\} $ is $i$-th domain binary label, $p(\widehat{y}_i; \bm{\theta})$ the probability of predicted label, and $\omega_0$, $\omega_1$
are constant weights. All experiments used $\omega_1 = 0.1$ and $\omega_0 = 0.9$, Adam optimizer~\cite{kingma2017adam} with the default parameters, Glorot normal
initialization~\cite{glorot} for weights, and zero initialization for biases.

Experiments were run at an Amazon EC2 instance with 36 CPU cores and 72GiB of RAM.

\subsection{Evaluation}%
\label{sub:evaluation}

The data (time windows) were split into train/validation/test sets as follows: weeks \texttt{06-10},
\texttt{06-27} and \texttt{07-22} were used for testing; week \texttt{06-24} for
validation, and all others for training. Hyperparameters were selected on the validation week to prevent overfitting.


The key component of the validation is the denylist which has two roles in the methodology: i) it is
used to compute features (Table~\ref{tab:features}) for model prediction, and ii) it is used as a
source of labels for evaluation of the prediction accuracy. Thus, it is essential that the
evaluation protocol prevents leaking of the labels to the features and, consequently, to predictions.
In the feature-generating role, the denylist is assumed to be incomplete, implying training the
model under the positive-unlabeled scenario. In the accuracy evaluation role, the denylist is
assumed to be complete, domains on the list are considered  malicious, and domains missing from
the list are considered benign.

These specifics are well known in the community, with available evaluation
protocols~\cite{Carter2014a,Jusko2017}, which we follow for comparison with the prior art. 
Specifically, the model is trained on all available training data, but the denylist corresponding to
the validation set is split randomly into $K$ disjoint folds. The evaluation runs for each fold
independently as follows: i) domains from the active fold are removed from the denylist (thus
considered `unknown'), and the features are computed only from denylists of the remaining folds; ii)
domains from the active fold and all benign domains are predicted by the model using the features.
After processing all $K$ folds, we have one prediction for all malicious samples and $K$ predictions
for the unknown samples. The final prediction of benign samples is obtained by aggregation using the
$\max$ operator to simulate the worst-case scenario with respect to false-positive rate. 


A potential drawback of using any evaluation that splits data into training and testing parts is
that the validated domain can be present in the training set with the correct label, hence the
learning system could have `memorized' it. We argue that this is highly unlikely since the used
features are in the form of statistics (Table~\ref{tab:features}). This conjecture is experimentally
validated using a sensitivity study (the Grill test) in Section~\ref{ssub:grill_test}.

\section{Experimental results}%
\label{sec:results}
Experiments are presented in four blocks. First, we show results when modeling domain relationship
with a single entity, which is needed for comparison to the prior art (PTP). Then, we present
experiments when all entities are used, highlighting the advantage of the proposed algorithm, since
PTP is not applicable. Following ablation studies investigate sensitivity to the size of the training
set and sensitivity to the noise in ground-truth labels.

Experimental results are presented as a precision-recall (PR) curve and a receiver operating
characteristic (ROC) curve with logarithmically scaled $ x $-axis to emphasize behavior on low false
positive rate. Average Precision (AP) and Area Under the ROC Curve (AUC) scalar metrics are used in
Table~\ref{tab:results_table} for summary. The main text contains results computed from one testing
week (\texttt{06-10}), the rest is available in the Supplementary material.

\subsection{Single relation (PTP comparison)}%
\begin{figure}
    \centering
    \begin{subfigure}[b]{0.49\columnwidth}
        \centering
        \includegraphics[width=\columnwidth]{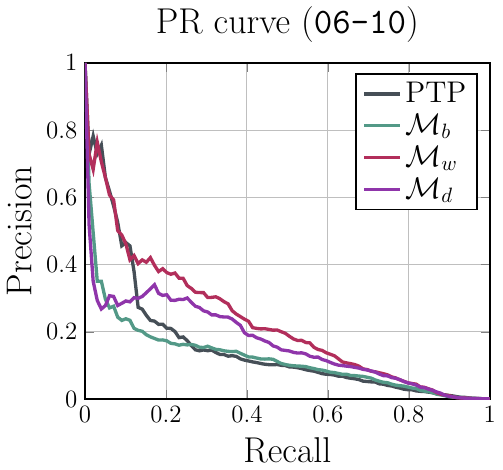}
    \end{subfigure}
    \hfill
    \begin{subfigure}[b]{0.49\columnwidth}
        \centering
        \includegraphics[width=\columnwidth]{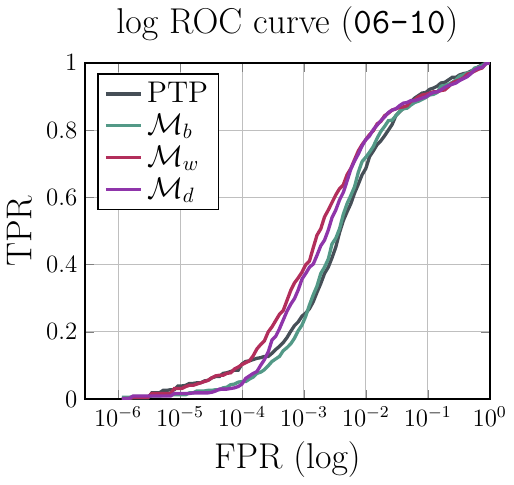}
    \end{subfigure}
    \caption{A PR curve and an ROC curve with logarithmic $ x $ axis comparing the performance of
    the three proposed architectures to the PTP algorithm. In these experiments, only one relation is employed.}
    \label{fig:ptpcomp_small}
\end{figure}

\label{ssub:comparison_to_ptp}
Probabilistic Threat Propagation (PTP) is compared to our algorithm on data sourced from  a
bipartite graph capturing \emph{domain-client} relations (recall that PTP cannot model multiple
relations). The implementation of PTP was taken from~\cite{Jusko2017} and used 20 iterations. From
PR and ROC curves shown in Figure~\ref{fig:ptpcomp_small} we see that both algorithms perform the
same for very low false positive rates (below $0.0001$) or low recall (below 0.15), but for higher
false positive rates and recall the proposed algorithm performs better. The results in the remaining
weeks from the testing set look similar and can be found in Supplementary material~\ref{ap:ptpcomp}.
The same graphs also show that wider networks $\mathcal{M}_w$ are better than the deeper network and
the baseline.


\subsection{All relations}%
\begin{figure}
    \centering
    \begin{subfigure}[b]{0.49\columnwidth}
        \centering
        \includegraphics[width=\textwidth]{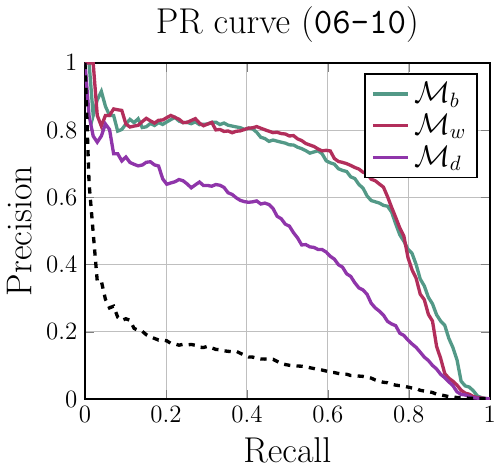}
    \end{subfigure}
    \hfill
    \begin{subfigure}[b]{0.49\columnwidth}
        \centering
        \includegraphics[width=\textwidth]{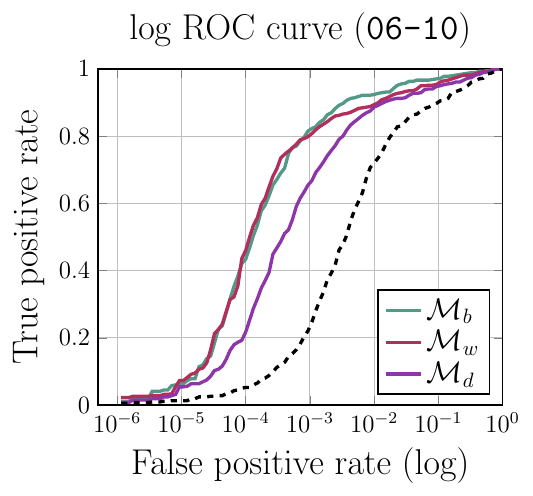}
    \end{subfigure}
    \caption{A PR curve and an ROC curve with logarithmic $ x $ axis comparing the performance of
    the three proposed architectures when all eleven relations are used. The black dashed curve
corresponds to the baseline ($ \mathcal{M}_b $) performance when only the domain-client relation is
used.}
    \label{fig:all_small}
\end{figure}

\label{ssub:all_relations}
Next, Figure~\ref{fig:all_small} (and Table~\ref{tab:results_table}) shows the PR and ROC curves
when the proposed method uses all eleven relations. For the sake of comparison, PTP is shown, but
keep in mind that PTP can use only the \emph{domain-client} graph due to its inherent limitations.
Adding additional relationships to the proposed model improves their performance approximately three
times according to average precision. Contrary to the above experiments, where the wider
architecture $ \mathcal{M}_w $ was clearly the best, here it performs similarly to  baseline
architecture $ \mathcal{M}_b.$  Deeper architecture $\mathcal{M}_d$ with a smaller number of neurons
was inferior. Results on the remaining weeks in the testing set are in Supplementary
material~\ref{ap:all_relations}.


\subsection{Size of the training set}%
\label{ssub:less_data}
The dependence of the proposed method on the size of the training set was studied by comparing
models training on graphs from the first 1, 3, 5, or 8 weeks (8 weeks means the full training set). The
training of all experiments used the same number of training steps (minibatches), potentially reusing
data more frequently if needed. The used experimental settings also allows the study of concept
drift, since small training sets used the oldest training set. The model used the baseline architecture
$ \mathcal{M}_b $ with all eleven relations.

Experimental results are summarised in Table~\ref{tab:results_table}. Surprisingly, there is a
relatively small dependence ($1-2$ percent) on the size of the training set. This is  probably
because the network traffic observed during different weeks is highly correlated, and the concept
drift present in the data  is not significant enough to alter the performance. Experimental results
on the remaining weeks are in Supplementary material~\ref{ap:results_lessgraphs}.


\subsection{Grill test}%
\label{ssub:grill_test}
To assess that the model does not overfit to specific domains seen during training, we used a
technique from \cite{Grill2016}, further called \emph{Grill test}. During the Grill test, we first
sample a subset of denylisted domains proportionally to the size of each malware family, omit them
from the sampled domains during training (they are never a central vertex), and if they appear in the
neighborhood of other domains, they are considered `unknown'. This way, the model cannot memorize
them as malware examples. During evaluation, only sampled domains are used as  ground truth for
positive labels. This tests the model on domains that have never been seen before and appear during
the testing first time.  Moreover, adding this noise to labels simulates incomplete denylist during
training, which surely happens in practice. Again a baseline architecture $ \mathcal{M}_b $ is
used for simplicity. 

According to the results shown in Table~\ref{tab:results_table}, there is a 3\% drop in performance
according to the AUC, which might be partially due to the noise but also due to the smaller number
of positive samples in the training set. AP has decreased by 30\%, which we attribute to the fact
that the PR curve is sensitive to class priors, which have changed. From these results we conclude that the
proposed method performs well on previously unseen malicious domains. Additional details can be
found in Supplementary material~\ref{ap:grill_test}.


\begin{table}
    \caption{Values of the AP and the AUC metrics for the three testing weeks. Each subtable
        represents a different experiment---\textit{single relation (comparison to PTP)}, \textit{all
        relations}, \textit{less training data}, and the \textit{Grill test}. The greatest score in each
        experiment and column is marked in bold. Note that the fifth and eleventh rows correspond to
        the identical settings, but are included for easier comparison.}%
    \label{tab:results_table}
    \centering
        {\renewcommand\arraystretch{1.2}
            \begin{tabular}{@{\hskip0pt}c@{\hskip5pt}c|cc|cc|cc}
                \toprule
            & & \multicolumn{2}{c|}{\texttt{06-10}} & \multicolumn{2}{c|}{\texttt{06-27}} & \multicolumn{2}{c}{\texttt{07-22}} \\
            & & AP & AUC & AP & AUC & AP & AUC \\
            \midrule
                \multirow{4}{*}{\begin{sideways}single rel.\end{sideways}} &
            PTP & $0.152$ & $\mathbf{0.962}$ & $0.120$ & $0.958$ & $0.142$ & $0.954$ \\
            & $ \mathcal{M}_b $ & $0.119$ & $0.958$ & $0.152$ & $0.959$ & $0.154$ & $0.955$ \\
            & $ \mathcal{M}_w $ & $\mathbf{0.221}$ & $ 0.956$ & $\mathbf{0.237}$ & $0.966$
                             & $\mathbf{0.220}$ & $0.957$ \\
            & $ \mathcal{M}_d $ & $0.163$ & $0.956$ & $0.189$ & $\mathbf{0.967}$ & $0.202$ &
            $\mathbf{0.958}$ \\
            \midrule
            \multirow{3}{*}{\begin{sideways}all rel.\end{sideways}} &
            $ \mathcal{M}_b $ & $0.637$ & $\mathbf{0.988}$ & $\mathbf{0.687}$ & $\mathbf{0.990}$ &
$\mathbf{0.657}$ & $\mathbf{0.988}$ \\
            & $ \mathcal{M}_w $ & $\mathbf{0.638}$ & $0.984$ & $ 0.678$ & $0.982$ & $0.654$ & $0.980$ \\
            & $ \mathcal{M}_d $ & $0.445$ & $0.978$ & $0.498$ & $0.983$ & $0.468$ & $0.980$ \\
            \midrule
            \multirow{4}{*}{\begin{sideways}less data\end{sideways}} &
            $1$ & $\mathbf{0.688}$ & $0.974$ & $0.696$ & $0.980$ &
            $0.665$ & $0.975$ \\
             & $3$ & $0.678$ & $0.983$ & $\mathbf{0.698}$ & $0.984$ &
            $\mathbf{0.670}$ & $0.981$ \\
             & $5$ & $0.633$ & $0.976$ & $0.671$ & $0.984$ & $0.631$ & $0.977$ \\
             & $8$ (all) & $0.637$ & $\mathbf{0.988}$ & $0.687$ &
            $\mathbf{0.990}$ & $0.657$ & $\mathbf{0.988}$ \\
            \midrule
            \begin{sideways} GT \end{sideways} & $ \mathcal{M}_b $ & $0.445$ & $0.951$ & $0.468$ & $0.966$ & $0.468$ & $0.959$ \\
            \bottomrule
            \end{tabular}
        }
\end{table}

\subsection{Practical discussion}%
\label{ssub:discussion}
The main aim of this work is to present a \emph{general} approach for malicious activity detection from interactions, which is not exclusive to only observing second-level domains. In a similar fashion, IP addresses, binary files, emails, or other network entities, could be used instead. As a result, even more possible attack surfaces would be monitored.

Since the cybersecurity landscape is rapidly evolving, HMILnet-based inference like most other methods leveraging statistical machine learning requires regular retraining. However, this is not difficult to perform---the method requires only ``raw'' bipartite graphs and updated denylist as input, no external features potentially hard or expensive to obtain. Moreover, our technique relies mostly on the behavior of the domains and not any other more specific features, which mitigates the need for frequent retraining.

The specific application of the proposed technique on second-level domain detection shares fundamental drawbacks and benefits with other detection engines that rely on domain reputation.
For example, otherwise legitimate domains are sometimes exploited for malicious activity. Domains suddenly acting maliciously would be detected by the method, as it is based on observing behavior directly. When the owner recovers the domain and it stops showing signs of malicious activity, the method will soon stop detecting it. The same reasoning could be for instance applied to fast-flux evasion techniques. In a graph that reflects domains' behavior, the properties of the immediate neighborhood of individual instances of a single fast-fluxed domain would be very similar. To avoid the detection, attackers would have to switch between individual instances frequently enough, or completely change the behavior from instance to instance. This increases attack costs and significantly lowers the appeal of such type of attack.

\section{Conclusion and Future work}%
\label{sec:conclusion}

This work demonstrated how the Hierarchical Multiple Instance Learning paradigm can be used to perform local inference in a streamlined neighborhood subgraph and compared it to the existing approaches. The HMILnet-based inference scales well to huge graphs with variable vertex degrees,  and due to the state heterogeneity property, it is possible to run inference on small subsets of vertices and the method is trivially parallelizable. We show that
HMILnet-based inference also performs well on the malicious domain discovery problem when only plain binary relations between entities in the network are employed. Our proposed method outperforms the state-of-the-art PTP algorithm when a PTP-compatible subset of data is used, however, the performance increases even more when all available data are used. Moreover, we demonstrate that the method produces a classifier that generalizes beyond simple memoization of known threats.

In comparison to the prior art PTP algorithm, where each edge can be described by, at most, a single scalar weight and the best choice is application-dependent, the proposed scheme handles rich hierarchical features on both vertices and edges. We consider attributing external features to vertices or edges an interesting future research direction as it will likely improve performance at the cost of increased computational complexity.

One more area that remains to be explored is to manually analyze some of the detected second-level domains and model behavior on them and then provide human-understandable interpretations for model decisions. 


\bibliography{library.bib} 
\bibliographystyle{ieeetr}

\clearpage

\begin{appendices}
\input{appendix_preprint}
\end{appendices}

\end{document}

%% file: figures/colors.tex
\usepackage{tikz}
\usepackage{xcolor}

\definecolor{domaincolor}{RGB}{178,46,91}
\definecolor{ipcolor}{RGB}{70,79,87}
\definecolor{emailcolor}{RGB}{241,156,56}
\definecolor{shacolor}{RGB}{143,53,170}
\definecolor{pathcolor}{RGB}{83,153,135}

\tikzstyle{edgestyle}=[thick, gray]
\tikzstyle{domainstyle}=[circle, minimum size=0.2cm, thick, draw=domaincolor, fill=domaincolor!85]
\tikzstyle{pathstyle}=[circle, minimum size=0.2cm, thick, draw=pathcolor, fill=pathcolor!85]
\tikzstyle{ipstyle}=[circle, minimum size=0.2cm, thick, draw=ipcolor, fill=ipcolor!85]
\tikzstyle{emailstyle}=[circle, minimum size=0.2cm, thick, draw=emailcolor, fill=emailcolor!85]
\tikzstyle{shastyle}=[circle, minimum size=0.2cm, thick, draw=shacolor, fill=shacolor!85]

%% file: appendix_preprint.tex
\section{Results: single relation}%
\label{ap:ptpcomp}

\begin{figure}[H]
    \centering
    \begin{subfigure}[b]{0.49\columnwidth}
        \centering
        \includegraphics[width=\textwidth]{figures/results/ptpcomp_pr0610_large.pdf}
    \end{subfigure}
    \hfill
    \begin{subfigure}[b]{0.49\columnwidth}
        \centering
        \includegraphics[width=\textwidth]{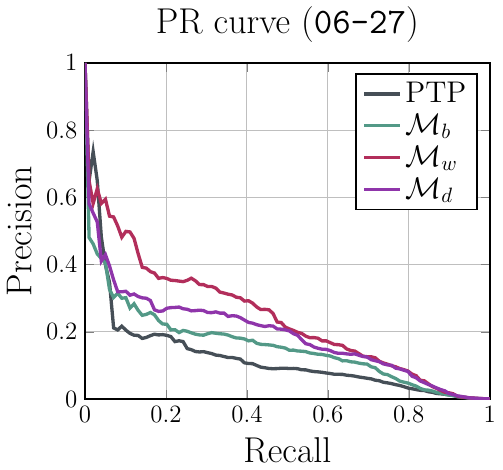}
    \end{subfigure}
    \hfill
    \begin{subfigure}[b]{0.49\columnwidth}
        \centering
        \includegraphics[width=\textwidth]{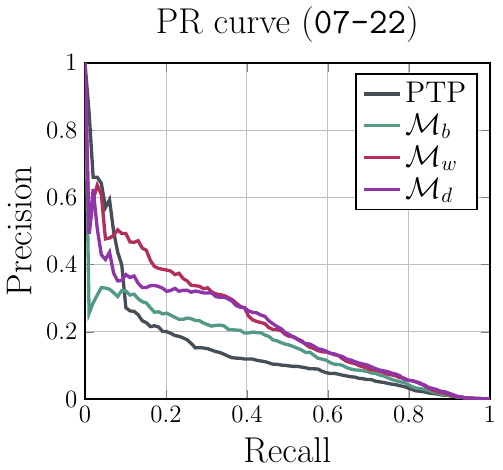}
    \end{subfigure}
    \caption{PR curves comparing the performance of the three proposed architectures (baseline $
    \mathcal{M}_b $, wider $ \mathcal{M}_w $, and deeper $ \mathcal{M}_d $) to the PTP algorithm.
Three figures are plotted, each corresponding to one of the testing datasets.}%
    \label{fig:ptpcomp_pr}
\end{figure}

\begin{figure}[H]
    \centering
    \begin{subfigure}[b]{0.49\columnwidth}
        \centering
        \includegraphics[width=\textwidth]{figures/results/ptpcomp_roc_log0610_large.pdf}
    \end{subfigure}
    \hfill
    \begin{subfigure}[b]{0.49\columnwidth}
        \centering
        \includegraphics[width=\textwidth]{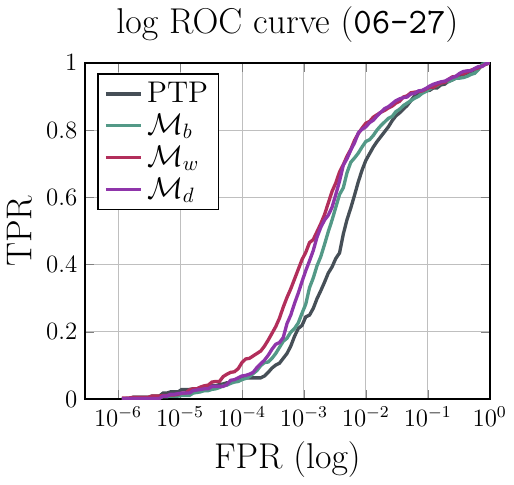}
    \end{subfigure}
    \hfill
    \begin{subfigure}[b]{0.49\columnwidth}
        \centering
        \includegraphics[width=\textwidth]{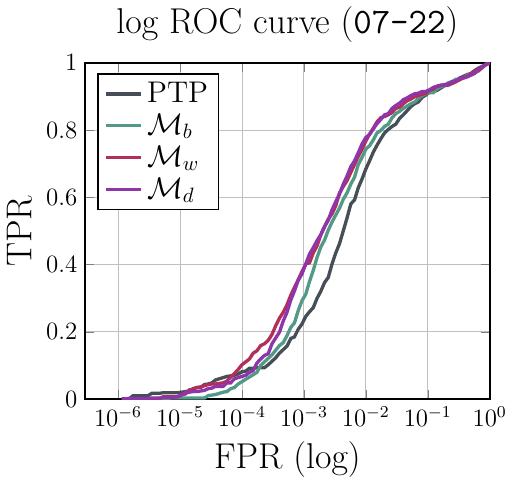}
    \end{subfigure}
    \caption{ROC curves comparing the performance of the three proposed architectures (baseline $
    \mathcal{M}_b $, wider $ \mathcal{M}_w $, and deeper $ \mathcal{M}_d $) to the PTP algorithm.
Three figures are plotted, each corresponding to one of the training datasets. The logarithmic scale
is used for $ x $-axis.}%
    \label{fig:ptpcomp_roc_log}
\end{figure}

\section{Results: all relations}%
\label{ap:all_relations}

\begin{figure}[H]
    \centering
    \begin{subfigure}[b]{0.49\columnwidth}
        \centering
        \includegraphics[width=\textwidth]{figures/results/all_pr0610_large.pdf}
    \end{subfigure}
    \hfill
    \begin{subfigure}[b]{0.49\columnwidth}
        \centering
        \includegraphics[width=\textwidth]{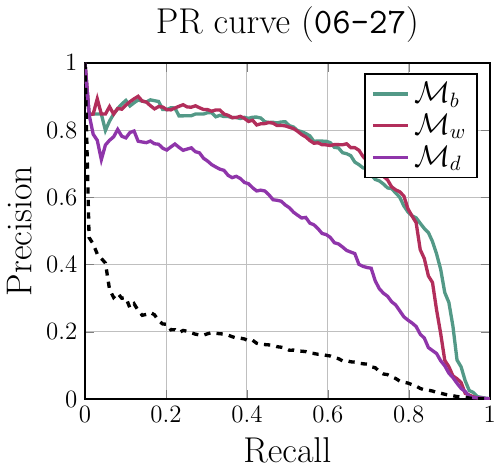}
    \end{subfigure}
    \hfill
    \begin{subfigure}[b]{0.49\columnwidth}
        \centering
        \includegraphics[width=\textwidth]{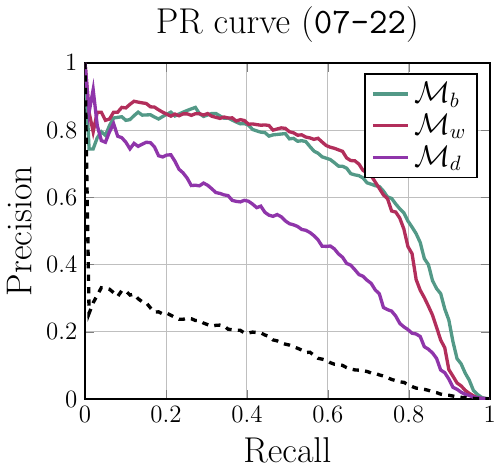}
    \end{subfigure}
    \caption{PR curves comparing the performance of three proposed architectures (baseline $
        \mathcal{M}_b $, wider $ \mathcal{M}_w $, and deeper $ \mathcal{M}_d $) when all eleven
        relations are used. Three figures are plotted, each corresponding to one of the testing
        datasets. The black dashed curve corresponds to the baseline ($ \mathcal{M}_b $) performance when
        only the domain-client relation is used.}%
    \label{fig:all_pr}
\end{figure}

\begin{figure}[H]
    \centering
    \begin{subfigure}[b]{0.49\columnwidth}
        \centering
        \includegraphics[width=\textwidth]{figures/results/all_roc_log0610_large.pdf}
    \end{subfigure}
    \hfill
    \begin{subfigure}[b]{0.49\columnwidth}
        \centering
        \includegraphics[width=\textwidth]{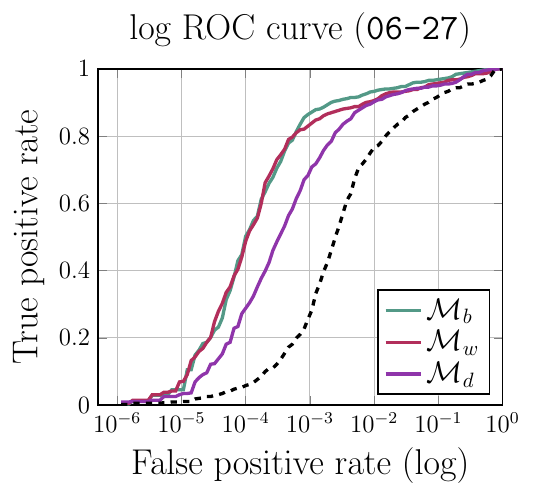}
    \end{subfigure}
    \hfill
    \begin{subfigure}[b]{0.49\columnwidth}
        \centering
        \includegraphics[width=\textwidth]{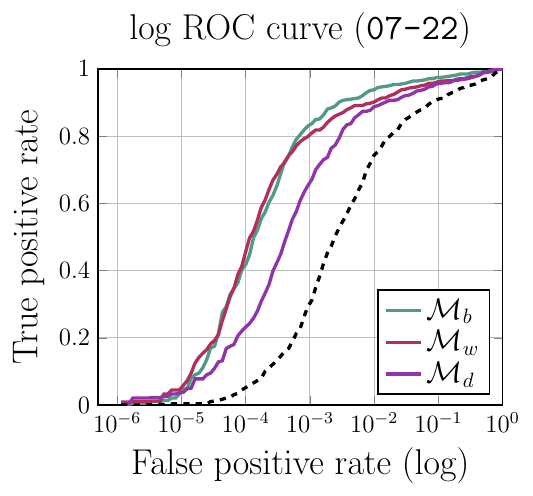}
    \end{subfigure}
    \caption{ROC curves comparing the performance of three proposed architectures (baseline $
    \mathcal{M}_b $, wider $ \mathcal{M}_w $, and deeper $ \mathcal{M}_d $) when all eleven
relations are used. Three figures are plotted, each corresponding to one of the training datasets. The
logarithmic scale is used for $ x $-axis. The black dashed curve corresponds to the baseline $
(\mathcal{M}_b $) performance when only the domain-client relation is used.}%
    \label{fig:all_roc_log}
\end{figure}

\section{Results: size of the training set}%
\label{ap:results_lessgraphs}

\begin{figure}[H]
    \centering
    \begin{subfigure}[b]{0.49\columnwidth}
        \centering
        \includegraphics[width=\textwidth]{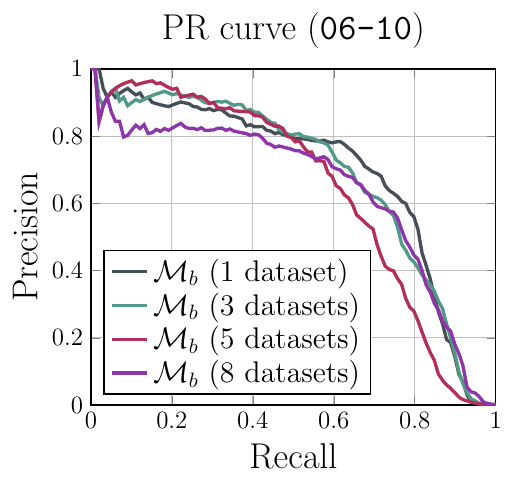}
    \end{subfigure}
    \hfill
    \begin{subfigure}[b]{0.49\columnwidth}
        \centering
        \includegraphics[width=\textwidth]{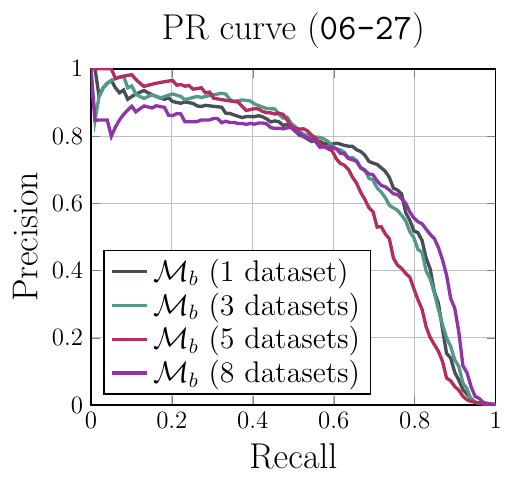}
    \end{subfigure}
    \hfill
    \begin{subfigure}[b]{0.49\columnwidth}
        \centering
        \includegraphics[width=\textwidth]{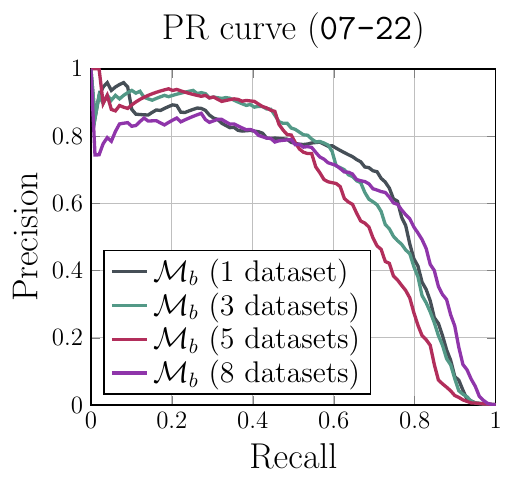}
    \end{subfigure}
    \caption{PR curves comparing the performance of the baseline architecture $ \mathcal{M}_b $ when a different number of datasets is used for training. Three figures are plotted, each corresponding to one of the testing datasets.}%
    \label{fig:lessgraphs_pr}
\end{figure}

\begin{figure}[H]
    \centering
    \begin{subfigure}[b]{0.49\columnwidth}
        \centering
        \includegraphics[width=\textwidth]{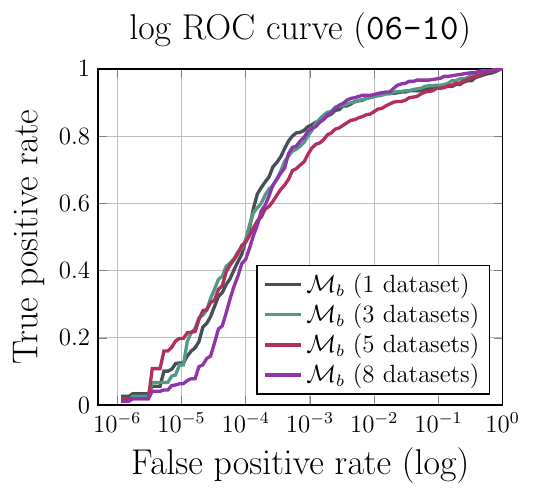}
    \end{subfigure}
    \hfill
    \begin{subfigure}[b]{0.49\columnwidth}
        \centering
        \includegraphics[width=\textwidth]{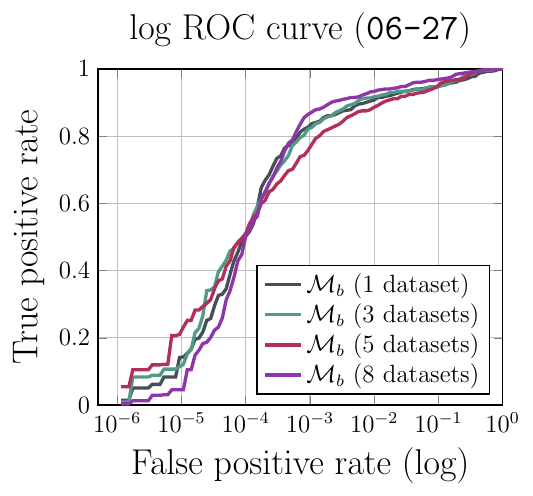}
    \end{subfigure}
    \hfill
    \begin{subfigure}[b]{0.49\columnwidth}
        \centering
        \includegraphics[width=\textwidth]{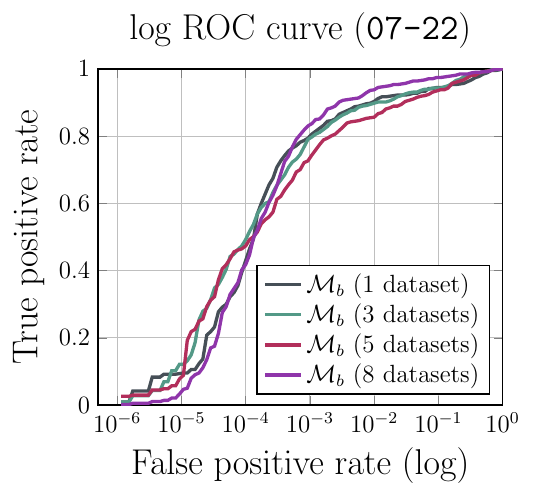}
    \end{subfigure}
    \caption{ROC curves comparing the performance of the baseline architecture $ \mathcal{M}_b $ when a different number of datasets is used for training. Three figures are plotted, each corresponding to one of the training datasets. The logarithmic scale is used for $ x $-axis.}%
    \label{fig:lessgraphs_roc}
\end{figure}

\section{Results: Grill test}%
\label{ap:grill_test}

\begin{figure}[H]
    \centering
    \begin{subfigure}[b]{0.49\columnwidth}
        \centering
        \includegraphics[width=\textwidth]{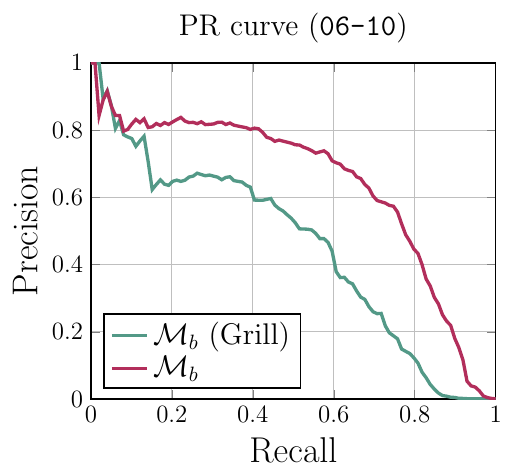}
    \end{subfigure}
    \hfill
    \begin{subfigure}[b]{0.49\columnwidth}
        \centering
        \includegraphics[width=\textwidth]{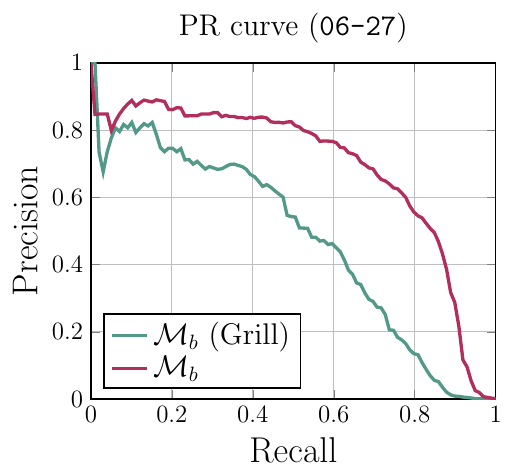}
    \end{subfigure}
    \hfill
    \begin{subfigure}[b]{0.49\columnwidth}
        \centering
        \includegraphics[width=\textwidth]{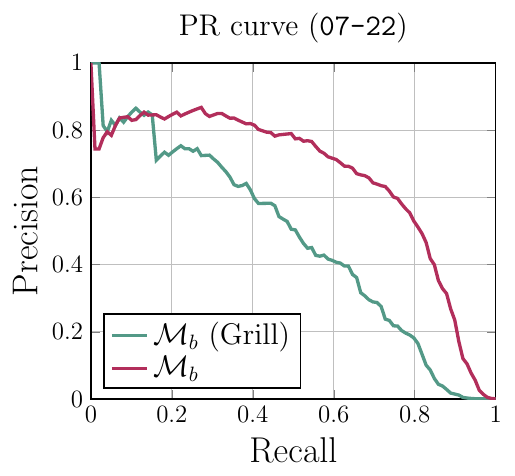}
    \end{subfigure}
    \caption{PR curves comparing the performance of the baseline architecture $ \mathcal{M}_b $ when the Grill test is (not) used. Three figures are plotted, each corresponding to one of the training datasets.}%
    \label{fig:gtest_pr}
\end{figure}

\begin{figure}[H]
    \centering
    \begin{subfigure}[b]{0.49\columnwidth}
        \centering
        \includegraphics[width=\textwidth]{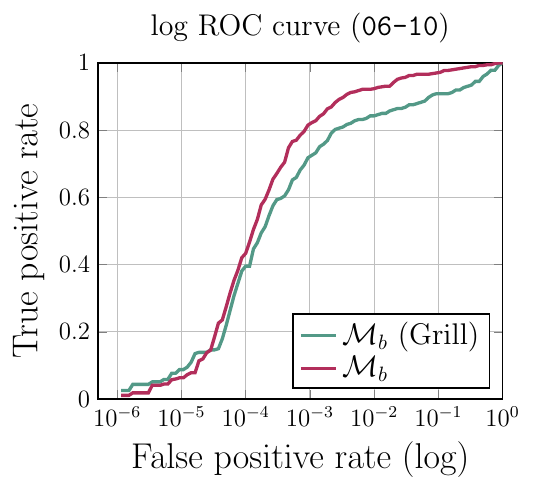}
    \end{subfigure}
    \hfill
    \begin{subfigure}[b]{0.49\columnwidth}
        \centering
        \includegraphics[width=\textwidth]{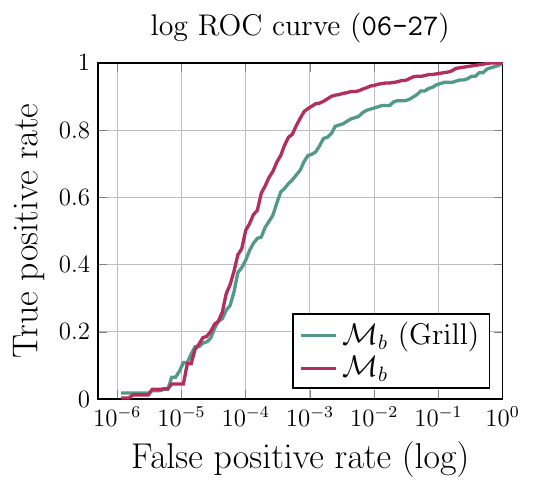}
    \end{subfigure}
    \hfill
    \begin{subfigure}[b]{0.49\columnwidth}
        \centering
        \includegraphics[width=\textwidth]{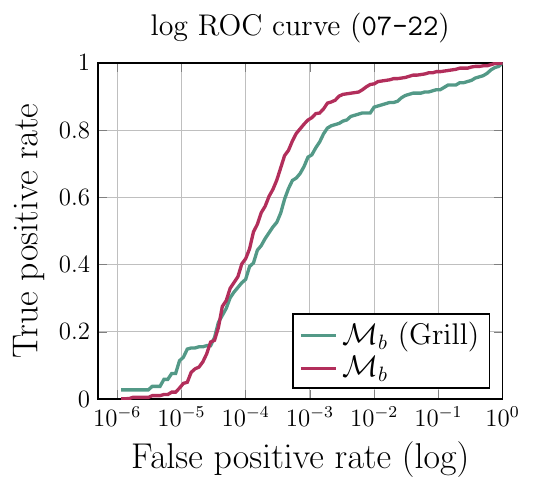}
    \end{subfigure}
    \caption{ROC curves comparing the performance of the baseline architecture $ \mathcal{M}_b $ when the Grill test is (not) used. Three figures are plotted, each corresponding to one of the training datasets. The logarithmic scale is used for $ x $-axis.}%
    \label{fig:gtest_roc_log}
\end{figure}

\onecolumn

\section{Data details}%
\label{ap:data_details}

\begin{table*}[!ht]
    \centering
    \caption{All relations used in our experiments, with their cardinality type and several examples
    of specific relation pairs (edges in a bipartite graph). Each edge is specified as $ (u; v) $ to
avoid confusion as commas may be used in $ u $ or  $ v $. M2M stands for `many-to-many' and M2O for
`many-to-one' cardinality type. Longer names are shortened using ellipsis (\ldots).}%
    \label{tab:relations}
    \renewcommand{\cellalign}{l}
    \resizebox{\textwidth}{!}{%
        \begin{tabular}{ccl}
            \toprule
            name (\texttt{domain}-*) & card. type & examples \\
            \midrule
            *-\texttt{client} & M2M & \makecell{(\texttt{tottenhamhotspur.com}; \texttt{S8g}) \\ (\texttt{loanstreet.com.my}; \texttt{2Pu3}) \\ (\texttt{healthlabtesting.com}; \texttt{2WLu})} \\
            \midrule
            *-\texttt{binary} & M2M & \makecell{  (\texttt{kotonoha-jiten.com}; \texttt{15CBF8\ldots}) \\ (\texttt{wonderslim.com}; \texttt{B41781\ldots}) \\ (\texttt{pythonprogramming.net}; \texttt{CF6ACB\ldots})}   \\
            \midrule
            *-\texttt{IP address} & M2M & \makecell{  (\texttt{quickpayportal.com};  \texttt{208.78.141.18}) \\ (\texttt{jobwinner.ch};  \texttt{217.71.91.48}) \\ (\texttt{tottenhamhotspur.com};  \texttt{104.16.54.111})  }\\
            \midrule
            *-\texttt{TLS issuer} & M2O & \makecell{ (\texttt{cratejoy.com};  \texttt{CN=Amazon, OU=Server CA 1B, O=Amazon, C=US}) \\ (\texttt{creative-serving.com};\\\quad  \texttt{CN=COMODO RSA Domain Validation Secure Server\ldots})\\ (\texttt{healthlabtesting.com}; \texttt{CN=Symantec Class 3 Secure Server CA\ldots})}\\
            \midrule
            *-\texttt{TLS hash} & M2O & \makecell{ (\texttt{timeoutdubai.com}; \texttt{90c093\ldots}) \\ (\texttt{boomerang.com}; \texttt{e577e6\ldots}) \\ (\texttt{quickpayportal.com}; \texttt{85bdd8\ldots})}    \\
            \midrule
            *-\texttt{TLS issue time} & M2O &  \makecell{(\texttt{jobwinner.ch}; \texttt{1496041693}) \\ (\texttt{healthlabtesting.com}; \texttt{1445558400}) \\ (\texttt{flatmates.com.au}; \texttt{1502150400})}   \\
            \midrule
            *-\texttt{WHOIS email} & M2O & \makecell{ (\texttt{unstableunicorns.com}; \texttt{unstableunicorns.com@*sbyproxy.com})\\ (\texttt{albertlee.biz}; \texttt{abuse@godaddy.com})\\  (\texttt{crowneplazalondonthecity.com};\\ \quad \texttt{crowneplazalondonthecity.com@*sbyproxy.com })  } \\
            \midrule
            *-\texttt{WHOIS nameserver} & M2O &  \makecell{(\texttt{grd779.com}; \texttt{ns2.hover.com}) \\ (\texttt{celeritascdn.com}; \texttt{lady.ns.cloudflare.com}) \\ (\texttt{smeresources.org}; \texttt{ns-495.awsdns-61.com})}   \\
            \midrule
            *-\texttt{WHOIS registrar name} & M2O & \makecell{ (\texttt{unblocked.how}; \texttt{eNom, Inc.}) \\ (\texttt{rev-stripe.com}; \texttt{Amazon Registrar, Inc.}) \\ (\texttt{chisaintjosephhealth.org}; \texttt{Register.com, Inc.})} \\
            \midrule
            *-\texttt{WHOIS country} & M2O & \makecell{ (\texttt{thefriscostl.com}; \texttt{CANADA}) \\ (\texttt{getwsone.com}; \texttt{UNITED STATES}) \\ (\texttt{notify.support}; \texttt{PANAMA})}\\
            \midrule
            *-\texttt{WHOIS registrar id} & M2O & \makecell{ (\texttt{lo3trk.com}; \texttt{468}) \\ (\texttt{watchcrichd.org}; \texttt{472}) \\ (\texttt{bozsh.com}; \texttt{1479}) }\\
            \midrule
            *-\texttt{WHOIS timestamp} & M2O & \makecell{ (\texttt{unblocked.how};  \texttt{15293}) \\ (\texttt{comicplanet.net}; \texttt{15159}) \\ (\texttt{unpublishedflight.com}; \texttt{15553})} \\
            \bottomrule
        \end{tabular}
    }
\end{table*}

\begin{table*}[!ht]
    \centering
    \caption{Numbers of vertices of each type in each dataset.}%
    \label{tab:sizes_nodes}
    \renewcommand{\cellalign}{cc}
    \resizebox{\textwidth}{!}{\begin{tabular}{r|*{13}{c}}
    \toprule
    date & \texttt{domain} & \texttt{client} & \texttt{binary} & \texttt{IP} & \makecell{\texttt{TLS} \\ \texttt{hash}} & \makecell{\texttt{TLS} \\ \texttt{issuer}} & \makecell{\texttt{TLS} \\ \texttt{issue} \\ \texttt{time}} & \makecell{\texttt{WHOIS} \\ \texttt{email}} & \makecell{\texttt{WHOIS} \\ \texttt{name-} \\ \texttt{server}} & \makecell{\texttt{WHOIS} \\ \texttt{registrar} \\ \texttt{name}} & \makecell{\texttt{WHOIS} \\ \texttt{country}} & \makecell{\texttt{WHOIS} \\ \texttt{registrar} \\ \texttt{id}} & \makecell{\texttt{WHOIS} \\ \texttt{time-} \\ \texttt{stamp}} \\
    \midrule
    \texttt{05-23} & $710\,167$ & $3\,419\,758$ & $245\,999$ & $2\,357\,441$ & $6\,217$ & $374\,531$ & $160\,343$ & $18\,537$ & $15\,972$ & $1\,111$ & $160$ & $959$ & $901$ \\
    \texttt{06-03} & $631\,828$ & $3\,151\,105$ & $232\,507$ & $2\,247\,940$ & $5\,631$ & $337\,558$ & $142\,944$ & $16\,442$ & $14\,803$ & $989$ & $157$ & $849$ & $901$ \\
    \texttt{06-10} & $644\,868$ & $3\,140\,231$ & $236\,052$ & $2\,268\,276$ & $5\,745$ & $348\,298$ & $148\,844$ & $16\,834$ & $14\,913$ & $969$ & $160$ & $833$ & $904$ \\
    \texttt{06-17} & $630\,532$ & $3\,114\,985$ & $242\,696$ & $2\,305\,323$ & $5\,618$ & $340\,934$ & $145\,032$ & $16\,808$ & $14\,973$ & $945$ & $161$ & $806$ & $904$ \\
    \texttt{06-26} & $648\,960$ & $3\,175\,500$ & $238\,535$ & $2\,316\,943$ & $5\,732$ & $350\,155$ & $149\,609$ & $17\,519$ & $15\,454$ & $954$ & $160$ & $805$ & $902$ \\
    \texttt{06-27} & $616\,958$ & $3\,106\,120$ & $240\,931$ & $2\,256\,445$ & $5\,484$ & $334\,996$ & $142\,453$ & $16\,468$ & $14\,942$ & $909$ & $158$ & $763$ & $901$ \\
    \texttt{07-01} & $613\,601$ & $3\,055\,840$ & $240\,909$ & $2\,277\,775$ & $5\,433$ & $332\,991$ & $141\,471$ & $16\,373$ & $14\,866$ & $894$ & $157$ & $743$ & $897$ \\
    \texttt{07-08} & $557\,253$ & $2\,894\,017$ & $224\,571$ & $2\,117\,016$ & $5\,077$ & $307\,541$ & $129\,249$ & $14\,319$ & $13\,756$ & $810$ & $160$ & $665$ & $890$ \\
    \texttt{07-15} & $608\,327$ & $2\,943\,900$ & $241\,631$ & $2\,222\,769$ & $5\,504$ & $330\,853$ & $140\,718$ & $16\,168$ & $14\,834$ & $861$ & $158$ & $717$ & $899$ \\
    \texttt{07-22} & $600\,814$ & $2\,929\,311$ & $239\,520$ & $2\,193\,113$ & $5\,423$ & $326\,592$ & $138\,604$ & $16\,214$ & $14\,764$ & $842$ & $156$ & $693$ & $898$ \\
    \texttt{07-29} & $588\,039$ & $2\,873\,682$ & $237\,436$ & $2\,162\,197$ & $5\,194$ & $320\,663$ & $135\,980$ & $15\,785$ & $14\,475$ & $806$ & $150$ & $656$ & $898$ \\
    \bottomrule
    \end{tabular}}
\end{table*}

\begin{table*}[!ht]
    \centering
    \caption{Numbers of edges between domains and other types of vertices in each dataset.}%
    \label{tab:sizes_edges}
    \renewcommand{\cellalign}{cc}
    \resizebox{\textwidth}{!}{\begin{tabular}{r|*{12}{c}}
    \toprule
    date & \texttt{client} & \texttt{binary} & \texttt{IP} & \makecell{\texttt{TLS} \\ \texttt{hash}} & \makecell{\texttt{TLS} \\ \texttt{issuer}} & \makecell{\texttt{TLS} \\ \texttt{issue} \\ \texttt{time}} & \makecell{\texttt{WHOIS} \\ \texttt{email}} & \makecell{\texttt{WHOIS} \\ \texttt{name-} \\ \texttt{server}} & \makecell{\texttt{WHOIS} \\ \texttt{registrar} \\ \texttt{name}} & \makecell{\texttt{WHOIS} \\ \texttt{country}} & \makecell{\texttt{WHOIS} \\ \texttt{registrar} \\ \texttt{id}} & \makecell{\texttt{WHOIS} \\ \texttt{time-} \\ \texttt{stamp}} \\
    \midrule
    \texttt{05-23} & $334\,320\,003$ & $4\,073\,050$ & $9\,314\,998$ & $465\,879$ & $465\,879$ & $465\,879$ & $44\,613$ & $122\,231$ & $42\,804$ & $39\,506$ & $46\,589$ & $46\,603$ \\
    \texttt{06-03} & $292\,659\,940$ & $3\,639\,635$ & $8\,356\,254$ & $419\,035$ & $419\,035$ & $419\,035$ & $38\,974$ & $108\,077$ & $37\,349$ & $34\,538$ & $40\,690$ & $40\,704$ \\
    \texttt{06-10} & $299\,532\,603$ & $3\,785\,134$ & $8\,902\,097$ & $430\,878$ & $430\,878$ & $430\,878$ & $39\,766$ & $110\,423$ & $38\,213$ & $35\,321$ & $41\,543$ & $41\,555$ \\
    \texttt{06-17} & $293\,466\,696$ & $3\,813\,715$ & $8\,803\,587$ & $420\,926$ & $420\,926$ & $420\,926$ & $39\,322$ & $109\,466$ & $37\,741$ & $34\,906$ & $41\,072$ & $41\,088$ \\
    \texttt{06-26} & $306\,114\,224$ & $3\,846\,992$ & $9\,141\,868$ & $431\,772$ & $431\,772$ & $431\,772$ & $41\,035$ & $114\,242$ & $39\,370$ & $36\,328$ & $42\,832$ & $42\,845$ \\
    \texttt{06-27} & $291\,065\,433$ & $3\,821\,056$ & $8\,740\,732$ & $412\,836$ & $412\,836$ & $412\,836$ & $38\,296$ & $107\,202$ & $36\,775$ & $33\,943$ & $39\,996$ & $40\,009$ \\
    \texttt{07-01} & $285\,920\,746$ & $3\,789\,277$ & $8\,649\,546$ & $410\,419$ & $410\,419$ & $410\,419$ & $37\,797$ & $106\,055$ & $36\,256$ & $33\,520$ & $39\,480$ & $39\,494$ \\
    \texttt{07-08} & $251\,616\,117$ & $3\,361\,849$ & $7\,917\,765$ & $376\,934$ & $376\,934$ & $376\,934$ & $33\,028$ & $936\,23$ & $317\,36$ & $292\,40$ & $345\,25$ & $345\,34$ \\
    \texttt{07-15} & $278\,227\,235$ & $3\,693\,630$ & $8\,869\,060$ & $406\,339$ & $406\,339$ & $406\,339$ & $37\,536$ & $105\,356$ & $36\,008$ & $33\,288$ & $39\,180$ & $39\,193$ \\
    \texttt{07-22} & $278\,975\,786$ & $3\,765\,391$ & $8\,627\,468$ & $400\,566$ & $400\,566$ & $400\,566$ & $37\,407$ & $104\,908$ & $35\,844$ & $33\,015$ & $39\,017$ & $39\,031$ \\
    \texttt{07-29} & $274\,507\,944$ & $3\,701\,055$ & $8\,475\,127$ & $392\,479$ & $392\,479$ & $392\,479$ & $36\,473$ & $102\,514$ & $35\,002$ & $32\,228$ & $38\,021$ & $38\,034$ \\
    \bottomrule
    \end{tabular}}
\end{table*}

\begin{table*}[!ht]
    \centering
    \caption{Numbers of malicious domains in each of the datasets. In the first row, there is the number of domains that are both in the denylist and in the observed dataset, and in the second row, we give the ratio of the number of denylisted domains to the total number of domains in the dataset.}%
    \label{tab:mal_numbers}
    \renewcommand{\cellalign}{l}
\resizebox{\textwidth}{!}{%
    \begin{tabular}{c|*{13}{c}}
    \toprule
    & \texttt{05-23} &\texttt{06-03} &\texttt{06-10} &\texttt{06-17} &\texttt{06-24}&\texttt{06-26} &\texttt{06-27} &\texttt{07-01} &\texttt{07-08} &\texttt{07-15} &\texttt{07-22} & \texttt{07-29} \\
            \midrule
        $ \lvert L \cap \mathcal{D} \rvert  $ & $656$ & $ 578 $ & $ 535 $ & $ 557 $& $ 538 $ & $ 557 $ & $ 552 $ & $ 555 $& $ 512 $& $ 557 $& $ 577 $& $ 548 $\\
        $ \lvert L \cap \mathcal{D} \rvert / {\lvert \mathcal{D} \rvert} $ & $ 0.924 $\textperthousand & $0.915$\textperthousand & $ 0.830 $\textperthousand & $ 0.883 $\textperthousand & $ 0.865 $\textperthousand& $ 0.858 $\textperthousand& $ 0.895 $\textperthousand& $ 0.904 $\textperthousand& $ 0.919 $\textperthousand& $ 0.916 $\textperthousand& $ 0.960 $\textperthousand & $ 0.932 $\textperthousand \\
    \bottomrule
    \end{tabular}}
\end{table*}